\documentclass[11pt,a4paper]{article}
\usepackage{times,latexsym}
\usepackage[hyphenbreaks]{breakurl}
\usepackage[hyphens]{url}
\usepackage[T1]{fontenc}

\usepackage[acceptedWithA,hyperref]{tacl2021v1}
\usepackage{booktabs}
\usepackage{verbatim}
\usepackage{amsmath}
\usepackage{array}
\usepackage{amsfonts}
\usepackage{bbm}
\usepackage{bm}
\usepackage{amssymb}
\usepackage{graphicx}
\usepackage{multirow}
\usepackage{amsthm}
\usepackage{float}
\usepackage{xcolor}
\usepackage{subcaption}
\usepackage{enumitem}

\usepackage{tikz}
\usetikzlibrary{math,calc,positioning,intersections,fit}
\usetikzlibrary{decorations.pathreplacing,decorations.markings,arrows.meta}
\tikzset{>=latex}

\usepackage{algorithm}
\usepackage[noend]{algpseudocode}
\usepackage{mathtools}
\usepackage{microtype}

\pgfdeclarelayer{bg}    
\pgfsetlayers{bg,main}  

\newtheorem{theorem}{Theorem}

\usepackage{amsmath,amsfonts,bm}

\def\eqref#1{equation~\ref{#1}}

\def\1{\bm{1}}

\def\vmu{{\bm{\mu}}}
\def\vtheta{{\bm{\theta}}}
\def\vpsi{{\bm{\psi}}}

\def\vb{{\bm{b}}}

\def\vd{{\bm{d}}}

\def\vm{{\bm{m}}}

\def\vp{{\bm{p}}}

\def\vt{{\bm{t}}}
\def\vu{{\bm{u}}}

\def\vw{{\bm{w}}}
\def\vx{{\bm{x}}}
\def\vy{{\bm{y}}}
\def\vz{{\bm{z}}}

\def\evpsi{{\psi}}

\def\evp{{p}}

\def\evw{{w}}
\def\evx{{x}}
\def\evy{{y}}

\def\mA{{\bm{A}}}

\def\mU{{\bm{U}}}

\DeclareMathAlphabet{\mathsfit}{\encodingdefault}{\sfdefault}{m}{sl}
\SetMathAlphabet{\mathsfit}{bold}{\encodingdefault}{\sfdefault}{bx}{n}

\newcommand{\E}{\mathbb{E}}

\newcommand{\R}{\mathbb{R}}

\DeclareMathOperator*{\argmax}{arg\,max}

\newcommand{\scan}{\textsc{Scan}}

\newcommand{\geoquery}{\textsc{GeoQuery}}
\newcommand{\clevr}{\textsc{Clevr}}
\newcommand{\closure}{\textsc{Closure}}
\newcommand{\iid}{\textsc{Iid}}
\newcommand{\rright}{\textsc{Right}}
\newcommand{\aright}{\textsc{ARight}}
\newcommand{\template}{\textsc{Template}}
\newcommand{\length}{\textsc{Length}}
\newcommand{\vphi}{\bm{\phi}}

\newcommand{\N}{\mathbb N}
\newcommand{\lang}{\mathcal G}
\newcommand{\tags}{E}
\newcommand{\types}{T}
\newcommand{\fargs}{f_{\textsc{Args}}}
\newcommand{\ftype}{f_{\textsc{Type}}}
\newcommand{\flabeling}{l}
\newcommand{\felabeling}{\bar l}

\DeclareMathOperator{\lmo}{lmo}

\newcommand{\indicator}{\delta}
\newcommand{\support}{\sigma}

\newcommand{\graph}{G}

\newcommand{\marginal}{\mathcal M}
\newcommand{\relmarginal}{\mathcal L}
\DeclareMathOperator{\conv}{conv}

\newcommand{\fullset}{\mathcal C}
\newcommand{\goldset}{\mathcal{C}^*}
\newcommand{\spanset}{\mathcal C^{\text{(sa)}}}
\newcommand{\valset}{\mathcal C^{\text{(val)}}}
\newcommand{\oneset}{\mathcal{C}^{\text{(one)}}}
\newcommand{\suploss}{\ell}
\newcommand{\weakloss}{\widetilde{\ell}}

\usepackage{xspace,mfirstuc,tabulary}

\newif\iftaclinstructions
\taclinstructionsfalse 
\iftaclinstructions
\renewcommand{\confidential}{}
\renewcommand{\anonsubtext}{(No author info supplied here, for consistency with
TACL-submission anonymization requirements)}
\newcommand{\instr}
\fi

\iftaclpubformat 

\else

\fi

\title{On graph-based reentrancy-free semantic parsing}

 \author{
   Alban Petit \and Caio Corro \\
   Universite Paris-Saclay, CNRS, LISN, 91400, Orsay, France \\
   \texttt{\{alban.petit,caio.corro\}@lisn.upsaclay.fr}
 }

\date{}

\begin{document}

\maketitle

\begin{abstract}
We propose a novel graph-based approach for semantic parsing that resolves two problems observed in the literature:
(1) seq2seq models fail on compositional generalization tasks;
(2) previous work using phrase structure parsers cannot cover all the semantic parses observed in treebanks.
We prove that both MAP inference and latent tag anchoring (required for weakly-supervised learning) are NP-hard problems.
We propose two optimization algorithms based on constraint smoothing and conditional gradient to approximately solve these inference problems.
Experimentally, our approach delivers state-of-the-art results on \geoquery{}, \scan{} and \clevr{}, both for i.i.d.\ splits and for splits that test for compositional generalization.
\end{abstract}

\let\svthefootnote\thefootnote

\let\thefootnote\relax\footnotetext{This work has been accepted for publication in TACL. This version is a pre-MIT Press publication version.}\let\thefootnote\svthefootnote

\section{Introduction}

Semantic parsing aims to transform a natural language utterance into a structured representation that can be easily manipulated by a software (for example to query a database).
As such, it is a central task in human-computer interfaces.
\citet{andreas-etal-2013-semantic} first proposed to rely on machine translation models for semantic parsing, where the target representation is linearized and treated as a foreign language.
Due to recent advances in deep learning and especially in sequence-to-sequence (seq2seq) with attention architectures for machine translation \cite{DBLP:journals/corr/BahdanauCB14}, it is appealing to use the same architectures for standard structured prediction problems \cite{vinyals-etal-2015-grammar}.
This approach is indeed common in semantic parsing \cite{jia-liang-2016-data, dong-lapata-2016-language, wang-etal-2020-rat}, \emph{inter alia}.
Unfortunately, there are well known limitations to seq2seq architectures for semantic parsing.
First, at test time, the decoding algorithm is typically based on beam search as the model is autoregressive and does not make any independence assumption.
In case of prediction failure, it is therefore unknown if this is due to errors in the weighting function or to the optimal solution failing out of the beam.
Secondly, they are known to fail when compositional generalization is required \cite{pmlr-v80-lake18a,finegan-dollak-etal-2018-improving,keysers2020measuring}.

In order to bypass these problems, \citet{herzig-berant-2021-span} proposed to represent the semantic content associated with an utterance as a phrase structure, \emph{i.e.}\ using the same representation usually associated with syntactic constituents.
As such, their semantic parser is based on standard span-based decoding algorithms \cite{hall-etal-2014-less,stern-etal-2017-minimal,corro-2020-span} with additional well-formedness constraints from the semantic formalism.
Given a weighting function, MAP inference is a polynomial time problem that can be solved via a variant of the CYK algorithm \cite{kasami1965efficient,YOUNGER1967189,cocke1969programming}.
Experimentally, \citet{herzig-berant-2021-span} show that their approach outperforms seq2seq models in terms of compositional generalization, therefore effectively bypassing the two major problems of these architectures.

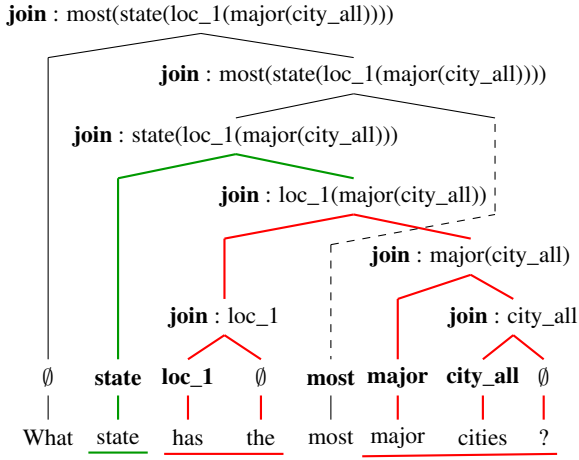
\begin{figure}
    \centering

\tikzmath{\x0 = 0; \x1 = 1.15; \x2 = 2.3; \x3 = 3.5; \x4 = 4.65; \x5 = 5.75; 
          \x6 = 7.15; \x7 = 8.15; \w1 = 0.3; \w2 = -0.3;
          \y0 = -3; \y1 = -2; \y2 = -1; \y3 = 0; \y4 = 1; \y5 = 2; \y6 = 3; \y7 = 4;
          \z6 = (\x6 + \x7) / 2; \z2 = (\x2 + \x3) / 2; \z5 = (\x5 + \x7) / 2;
          \z7 = (\z2 + \x6) / 2; \z4 = (\x1 + \z7) / 2; \z0 = (\x0 + \z7) / 2; }

\begin{tikzpicture}[scale=0.80,every node/.style={scale=0.80}]
    
    \node[] (what) at (\x0, \y0) {What};
    \node[] (state) at (\x1, \y0) {state};
    \node[] (has) at (\x2, \y0) {has};
    \node[] (the) at (\x3, \y0) {the};
    \node[] (most) at (\x4, \y0) {most};
    \node[] (major) at (\x5, \y0) {major};
    \node[] (cities) at (\x6, \y0) {cities};
    \node[] (q) at (\x7, \y0) {?};
    
    \draw[black!40!green, thick] (state.south west) -- (state.south east);
    \draw[red, thick] (has.south west) -- (the.south east);
    \draw[red, thick] (major.south west) -- (q.south east);
    
    \node[] (empty0) at (\x0, \y1) {$\emptyset$};
    \node[] (state_p) at (\x1, \y1) {\textbf{state}};
    \node[] (loc_1) at (\x2, \y1) {\textbf{loc\_1}};
    \node[] (empty3) at (\x3, \y1) {$\emptyset$};
    \node[] (most_p) at (\x4, \y1) {\textbf{most}};
    \node[] (major_p) at (\x5, \y1) {\textbf{major}};
    \node[] (city_all) at (\x6, \y1) {\textbf{city\_all}};
    \node[] (empty7) at (\x7, \y1) {$\emptyset$};
    
    \node[] (join2-3) at (\z2, \y2) {\textbf{join} : loc\_1};
    \node[] (join6-7) at (\z6, \y2) {\textbf{join} : city\_all};
    \node[] (join5-7) at (\z5, \y3) {\textbf{join} : major(city\_all)};
    \node[] (join2-7) at (\z7, \y4) {\textbf{join} : loc\_1(major(city\_all))};
    \node[] (join1-7) at (\z4, \y5) {\textbf{join} : state(loc\_1(major(city\_all)))};
    \node[] (bridge) at (\z7, \y6) {\textbf{join} : most(state(loc\_1(major(city\_all))))};
    \node[] (join0-7) at (\z0, \y7) {\textbf{join} : most(state(loc\_1(major(city\_all))))};
    
    \draw ($(what.center)+(0,\w1)$) to ($(empty0.center)+(0,\w2)$);
    \draw[black!40!green,thick] ($(state.center)+(0,\w1)$) to ($(state_p.center)+(0,\w2)$);
    \draw[red,thick] ($(has.center)+(0,\w1)$) to ($(loc_1.center)+(0,\w2)$);
    \draw[red,thick] ($(the.center)+(0,\w1)$) to ($(empty3.center)+(0,\w2)$);
    \draw ($(most.center)+(0,\w1)$) to ($(most_p.center)+(0,\w2)$);
    \draw[red,thick] ($(major.center)+(0,\w1)$) to ($(major_p.center)+(0,\w2)$);
    \draw[red,thick] ($(cities.center)+(0,\w1)$) to ($(city_all.center)+(0,\w2)$);
    \draw[red,thick] ($(q.center)+(0,\w1)$) to ($(empty7.center)+(0,\w2)$);
    \draw[red,thick] ($(join2-3.center)+(0,\w2)$) to ($(loc_1.center)+(0,\w1)$);
    \draw[red,thick] ($(join2-3.center)+(0,\w2)$) to ($(empty3.center)+(0,\w1)$);
    \draw[red,thick] ($(join6-7.center)+(0,\w2)$) to ($(city_all.center)+(0,\w1)$);
    \draw[red,thick] ($(join6-7.center)+(0,\w2)$) to ($(empty7.center)+(0,\w1)$);
    \draw[red,thick] ($(join5-7.center)+(0,\w2)$) to ($(join6-7.center)+(0,\w1)$);
    \draw[red,thick] ($(join5-7.center)+(0,\w2)$) to ($(major_p.center)+(0,1.3)$);
    \draw[red,thick] ($(major_p.center)+(0,\w1)$) to ($(major_p.center)+(0,1.3)$);
    \draw[red,thick] ($(join5-7.center)+(0,\w1)$) to ($(join2-7.center)+(0,\w2)$);
    \draw[red,thick] ($(join2-7.center)+(0,\w2)$) to ($(join2-3.center)+(0,1.3)$);
    \draw[red,thick] ($(join2-3.center)+(0,\w1)$) to ($(join2-3.center)+(0,1.3)$);
    \draw[black!40!green,thick] ($(join2-7.center)+(0,\w1)$) to ($(join1-7.center)+(0,\w2)$);
    \draw[black!40!green,thick] ($(join1-7.center)+(0,\w2)$) to ($(state_p.center)+(0,3.3)$);
    \draw[black!40!green,thick] ($(state_p.center)+(0,\w1)$) to ($(state_p.center)+(0,3.3)$);
    \draw ($(bridge.center)+(0,\w1)$) to ($(join0-7.center)+(0,\w2)$);
    \draw ($(join0-7.center)+(0,\w2)$) to ($(empty0.center)+(0,5.3)$);
    \draw ($(empty0.center)+(0,\w1)$) to ($(empty0.center)+(0,5.3)$);
    
    \draw ($(bridge.center)+(0,\w2)$) to ($(join1-7.center)+(0,\w1)$);
    \draw[dashed] ($(most_p.center)+(0,2.2)$) to ($(most_p.center)+(0,\w1)$);
    \draw[dashed] ($(most_p.center)+(0,2.2)$) to ($(most_p.center)+(2.7,2.7)$);
    \draw ($(bridge.center)+(0,\w2)$) to ($(most_p.center)+(2.7,4.3)$);
    \draw[dashed] ($(most_p.center)+(2.7,4.2)$) to ($(most_p.center)+(2.7,2.7)$);

\end{tikzpicture}
    \caption{Example of a semantic phrase structure from \geoquery{}. This structure is outside of the search space of the parser of \citet{herzig-berant-2021-span} as the constituent in red is discontinuous and also has a discontinuous parent (in red+green).}
    \label{fig:span}
\end{figure}

The complexity of MAP inference for phrase structure parsing is directly impacted by the search space considered \cite{kallmeyer2010parsing}.
Importantly, (ill-nested) discontinuous phrase structure parsing is known to be NP-hard, even with a bounded block-degree \cite{satta-1992-recognition}.
\citet{herzig-berant-2021-span} explore two restricted inference algorithms, both of which have a cubic time complexity with respect to the input length.
The first one only considers continuous phrase structures, \emph{i.e.}\,derived trees that could have been generated by a context-free grammar, and the second one also considers a specific type of discontinuities, see \citet[][Section 3.6]{corro-2020-span}.
Both algorithms fail to cover the full set of phrase structures observed in semantic treebanks, see Figure~\ref{fig:span}.

In this work, we propose to reduce semantic parsing without reentrancy (\emph{i.e.}\ a given predicate or entity cannot be used as an argument for two different predicates) to a bi-lexical dependency parsing problem.
As such, we tackle the same semantic content as aforementioned previous work but using a different mathematical representation \cite{rambow-2010-simple}.
We identify two main benefits to our approach:
(1) as we allow crossing arcs, \emph{i.e.}\ ``non-projective graphs'', all datasets are guaranteed to be fully covered
and
(2) it allows us to rely on optimization methods to tackle inference intractability of our novel graph-based formulation of the problem.
More specifically, in our setting we need to jointly assign predicates/entities to words that convey a semantic content and to identify arguments of predicates via bi-lexical dependencies.
We show that MAP inference in this setting is equivalent to the maximum generalized spanning arborescence problem \cite{myung1995gmsa} with supplementary constraints to ensure well-formedness with respect to the semantic formalism.
Although this problem is NP-hard, we propose an optimization algorithm that solves a linear relaxation of the problem and can deliver an optimality certificate.

Our contributions can be summarized as follows:
\begin{itemize}
    \item We propose a novel graph-based approach for semantic parsing without reentrancy;
    \item We prove the NP-hardness of MAP inference and latent anchoring inference;
	\item We propose a novel integer linear programming formulation for this problem together with an approximate solver based on conditional gradient and constraint smoothing;
	\item We tackle the training problem using variational approximations of objective functions, including the weakly-supervised scenario;
	\item We evaluate our approach on \geoquery{}, \scan{} and \clevr{} and observe that it outperforms baselines on both i.i.d.\ splits and splits that test for compositional generalization.
\end{itemize}
Code to reproduce the experiments is available online.\footnote{\url{https://github.com/alban-petit/semantic-dependency-parser}}
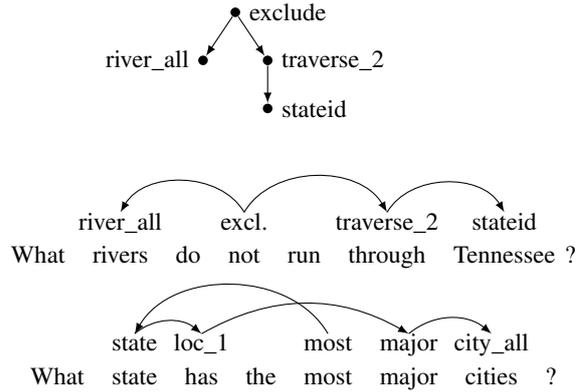
\begin{figure}
    \begin{center}

%
%

%
%
\begin{tikzpicture}[
    scale=0.85,
    every node/.style={
        scale=0.85
    }
]
    \node[scale=1.03, fill=white, inner sep=0pt, rectangle] at (0, 0.75) {exclude ( river\_all , traverse\_2 ( stateid ) )};
\end{tikzpicture}
\newline\newline
%
%
\begin{tikzpicture}[
    scale=0.85,
    every node/.style={
        scale=0.85
    },
    ast/.style={
        circle,
        fill=black,
        inner sep=1.5pt
    }
]
    \node[ast,label={right:exclude}] (exclude) at (0, -0.75) {};
    \node[ast,label={left:river\_all}] (riverall) at (-0.5, -1.5) {};
    \node[ast,label={right:traverse\_2}] (traverse2) at (0.5, -1.5) {};
    \node[ast,label={right:stateid}] (stateid) at (0.5, -2.25) {};
    
    \draw[->] (exclude) -- (riverall);
    \draw[->] (exclude) -- (traverse2);
    \draw[->] (traverse2) -- (stateid);
\end{tikzpicture}
\newline\newline
%
%
\begin{tikzpicture}[
    scale=0.85,
    every node/.style={
        scale=0.85
    },
    word/.style={
        rectangle,
        inner sep=0pt,
        node distance=10pt,
        text height=1.5ex,
        text depth=.25ex
    },
    tag/.style={
        rectangle,
        inner sep=0pt,
        node distance=5pt,
        text height=1.5ex,
        text depth=.25ex
    }
]
    \node[word] (what) {What};
    
    \node[word, right=of what] (rivers) {rivers};
    \node[tag, above=of rivers] (riverall) {river\_all};
    
    \node[word, right=of rivers] (do) {do};
    
    \node[word, right=of do] (not) {not};
    \node[tag, above=of not] (exclude) {excl.};
    
    \node[word, right=of not] (run) {run};
    
    \node[word, right=of run] (through) {through};
    \node[tag, above=of through] (traverse2) {traverse\_2};
    
    \node[word, right=of through] (Tennessee) {Tennessee};
    \node[tag, above=of Tennessee] (stateid) {stateid};
    
    \node[word, right=of Tennessee,xshift=-8pt] (mark) {?};

    \draw[->] (exclude.north) to[bend right=60] (riverall.north);
    \draw[->] (exclude.north) to[bend left=60] (traverse2.north);
    \draw[->] (traverse2.north) to[bend left=60] (stateid.north);
\end{tikzpicture}
\begin{tikzpicture}[
    scale=0.85,
    every node/.style={
        scale=0.85
    },
    word/.style={
        rectangle,
        inner sep=0pt,
        node distance=10pt,
        text height=1.5ex,
        text depth=.25ex
    },
    tag/.style={
        rectangle,
        inner sep=0pt,
        node distance=5pt,
        text height=1.5ex,
        text depth=.25ex
    }
]
    \node[word] (what) {What};
    
    \node[word, right=of what] (state) {state};
    \node[tag, above=of state] (state2) {state};
    
    \node[word, right=of state] (has) {has};
    \node[tag, above=of has] (loc1) {loc\_1};
    
    \node[word, right=of has] (the) {the};
    
    \node[word, right=of the] (most) {most};
    \node[tag, above=of most] (most2) {most};
    
    \node[word, right=of most] (major) {major};
    \node[tag, above=of major] (major2) {major};
    
    \node[word, right=of major] (cities) {cities};
    \node[tag, above=of cities] (cityall) {city\_all};
    
    \node[word, right=of cities] (mark) {?};

    \draw[->] (most2.north) to[bend right=60] (state2.north);
    \draw[->] (state2.north)  to[bend left=40] (loc1.north);
    \draw[->] (loc1.north)  to[bend left=30] (major2.north);
    \draw[->] (major2.north)  to[bend left=40] (cityall.north);
\end{tikzpicture}
    \end{center}
    \caption{%
    \textbf{(top)} The semantic program corresponding to the sentence ``What rivers do not run through Tennessee?'' in the \geoquery{} dataset.
    \textbf{(middle)} The associated AST.
    \textbf{(bottom)} Two examples illustrating the intuition of our model: we jointly assign predicates/entities and identify argument dependencies. As such, the resulting structure is strongly related to a syntactic dependency parse, but where the dependency structure do not cover all words.
    }
    \label{fig:ast}
\end{figure}
\section{Graph-based semantic parsing}
\begin{figure*}[!ht]
    \centering
    \begin{subfigure}{0.475\textwidth}
        \centering
        \tikzstyle{word} = [rectangle, rounded corners, text centered, minimum height=1em, anchor=mid]
\tikzstyle{tree_node} = [scale=0.8, rectangle, text centered, minimum height=1.2em, anchor=mid, draw=black]
\tikzmath{\x0 = 0; \x1 = 1.45; \x2 = 2.75; \x3 = 3.75; \x4 = 5.05; \x5 = 6.5; \x6=7.5; }
\tikzstyle{arrow} = [thick,->,>=stealth]

\definecolor{LightGray}{rgb}{0.95, 0.95, 0.95}

\begin{tikzpicture}[
    scale=0.8,
    every node/.style={scale=0.8},
    sem_node/.style={
        scale=0.85,
        circle,
        fill=black,
        inner sep=1.5pt
    }
]
\node (word_1) at (\x0, -4) [word] {Which};
\node (word_2) at (\x1, -4) [word] {states};
\node (word_3) at (\x2, -4) [word] {do};
\node (word_4) at (\x3, -4) [word] {not};
\node (word_5) at (\x4, -4) [word] {border};
\node (word_6) at (\x5, -4) [word] {texas};
\node (word_8) at (\x6, -4) [word] {?};
\phantom{
\node[label={[name=fake_node_label2]left:state\_all}] (fake_node) at (\x1-0.5, -4) [sem_node] {};
\node[label={[name=fake_node_label3]right:stateid}] (stateid) at (\x5, -4) [sem_node] {};
}
\begin{pgfonlayer}{bg}
    \draw[rounded corners,fill=LightGray,LightGray]
        ($(fake_node_label2.north west)+(-0.15,0.15)$)
        rectangle
        ($(fake_node_label3.south east)+(0.15,-0.15)$)
    ;
\end{pgfonlayer}

\node[label={right:exclude}] (exclude) at (\x3, 0) [sem_node] {};
\node[label={right:next\_to\_2}] (nextto2) at (\x4, -1) [sem_node] {};
\node[label={[name=stateid_label]right:stateid}] (stateid) at (\x5, -2) [sem_node] {};
\node[label={left:state\_all}] (stateall) at (\x1, -1) [sem_node] {};

\draw[->] (exclude) -- (nextto2);
\draw[->] (nextto2) -- (stateid);
\draw[->] (exclude) -- (stateall);

\phantom{
\node[label={[name=fake_node_label]left:state\_all}] (fake_node) at (\x1-0.5, 0) [sem_node] {};
}
\begin{pgfonlayer}{bg}
    \draw[rounded corners,fill=LightGray,LightGray]
        ($(fake_node_label.north west)+(-0.15,0.15)$)
        rectangle
        ($(stateid_label.south east)+(0.15,-0.15)$)
    ;
\end{pgfonlayer}

\draw [dashed] (exclude) -- (word_4);
\draw [dashed] (nextto2) -- (word_5);
\draw [dashed] (stateid) -- (word_6);
\draw [dashed] (stateall) -- (word_2);

\end{tikzpicture}

        \caption{}
        \label{fig:dependency}
    \end{subfigure}%
    \hfill%
    \begin{subfigure}{0.525\textwidth}
        \centering%
            \definecolor{Red}{rgb}{1, 0, 0}
\definecolor{Gray}{rgb}{0.5, 0.5, 0.5}
\definecolor{BurntOrange}{rgb}{1, 0.5, 0}

\tikzmath{\x0 = 0; \x1 = 1.25; \x2 = 2.35; \x3 = 3.45; \x4 = 4.6; \x5 = 5.75; \x6=6.9; \x7=7.5; }

\tikzstyle{word} = [rectangle, rounded corners, text width=5.5em, text centered, minimum height=2em, anchor=mid]

\begin{tikzpicture}[scale=0.8,every node/.style={scale=0.8}]
    \node[circle,fill=Red,inner sep=1.5pt] (root) at (\x0-0.85, 1) {};
    
    \draw[thick,dashed,rounded corners,Gray]
        ($(root.north west)+(-0.15,0.15)$)
        rectangle
        ($(root.south east)+(0.15,-0.15)$)
    ;
    
    \node[circle,fill=BurntOrange,inner sep=1.5pt] (which_null) at (\x0, 0) {};
    \node[circle,fill=black,inner sep=1.5pt] at (\x0, -0.5) {};
    \node[circle,fill=black,inner sep=1.5pt] at (\x0, -1) {};
    \node[circle,fill=black,inner sep=1.5pt] at (\x0, -1.5) {};
    \node[circle,fill=black,inner sep=1.5pt] at (\x0, -2) {};
    \node[circle,fill=black,inner sep=1.5pt] (which_state_all) at (\x0, -2.5) {};
    \node[word] at (\x0, -3.2) {Which};
    \draw[thick,dashed,rounded corners,Gray]
        ($(which_null.north west)+(-0.15,0.15)$)
        rectangle
        ($(which_state_all.south east)+(0.15,-0.15)$)
    ;
    
    \node[circle,fill=black,inner sep=1.5pt] (states_null) at (\x1, 0) {};
    \node[circle,fill=black,inner sep=1.5pt] at (\x1, -0.5) {};
    \node[circle,fill=black,inner sep=1.5pt] at (\x1, -1) {};
    \node[circle,fill=black,inner sep=1.5pt] at (\x1, -1.5) {};
    \node[circle,fill=Red,inner sep=1.5pt] (states_state_all) at (\x1, -2) {};
    \node[circle,fill=black,inner sep=1.5pt] (states_area) at (\x1, -2.5) {};
    \node[word] at (\x1, -3.2) {states};
    \draw[thick,dashed,rounded corners,Gray]
        ($(states_null.north west)+(-0.15,0.15)$)
        rectangle
        ($(states_area.south east)+(0.15,-0.15)$)
    ;
    
    \node[circle,fill=BurntOrange,inner sep=1.5pt] (do_null) at (\x2, 0) {};
    \node[circle,fill=black,inner sep=1.5pt] at (\x2, -0.5) {};
    \node[circle,fill=black,inner sep=1.5pt] at (\x2, -1) {};
    \node[circle,fill=black,inner sep=1.5pt] at (\x2, -1.5) {};
    \node[circle,fill=black,inner sep=1.5pt] at (\x2, -2) {};
    \node[circle,fill=black,inner sep=1.5pt] (do_state_all) at (\x2, -2.5) {};
    \node[word] at (\x2, -3.2) {do};
    \draw[thick,dashed,rounded corners,Gray]
        ($(do_null.north west)+(-0.15,0.15)$)
        rectangle
        ($(do_state_all.south east)+(0.15,-0.15)$)
    ;
    
    \node[circle,fill=black,inner sep=1.5pt] (not_null) at (\x3, 0) {};
    \node[circle,fill=Red,inner sep=1.5pt] (not_exclude) at (\x3, -0.5) {};
    \node[circle,fill=black,inner sep=1.5pt] at (\x3, -1) {};
    \node[circle,fill=black,inner sep=1.5pt] at (\x3, -1.5) {};
    \node[circle,fill=black,inner sep=1.5pt] at (\x3, -2) {};
    \node[circle,fill=black,inner sep=1.5pt] (not_state_all) at (\x3, -2.5) {};
    \node[word] at (\x3, -3.2) {not};
    \draw[thick,dashed,rounded corners,Gray]
        ($(not_null.north west)+(-0.15,0.15)$)
        rectangle
        ($(not_state_all.south east)+(0.15,-0.15)$)
    ;
    
    \node[circle,fill=black,inner sep=1.5pt] (border_null) at (\x4, 0) {};
    \node[circle,fill=black,inner sep=1.5pt] at (\x4, -0.5) {};
    \node[circle,fill=Red,inner sep=1.5pt] (border_border) at (\x4, -1) {};
    \node[circle,fill=black,inner sep=1.5pt] at (\x4, -1.5) {};
    \node[circle,fill=black,inner sep=1.5pt] at (\x4, -2) {};
    \node[circle,fill=black,inner sep=1.5pt] (border_state_all) at (\x4, -2.5) {};
    \node[word] at (\x4, -3.2) {border};
    \draw[thick,dashed,rounded corners,Gray]
        ($(border_null.north west)+(-0.15,0.15)$)
        rectangle
        ($(border_state_all.south east)+(0.15,-0.15)$)
    ;
    
    \node[circle,fill=black,inner sep=1.5pt] (texas_null) at (\x5, 0) {};
    \node[circle,fill=black,inner sep=1.5pt] at (\x5, -0.5) {};
    \node[circle,fill=black,inner sep=1.5pt] at (\x5, -1) {};
    \node[circle,fill=Red,inner sep=1.5pt] (texas_stateid) at (\x5, -1.5) {};
    \node[circle,fill=black,inner sep=1.5pt] at (\x5, -2) {};
    \node[circle,fill=black,inner sep=1.5pt] (texas_state_all) at (\x5, -2.5) {};
    \node[word] at (\x5, -3.2) {texas};
    \draw[thick,dashed,rounded corners,Gray]
        ($(texas_null.north west)+(-0.15,0.15)$)
        rectangle
        ($(texas_state_all.south east)+(0.15,-0.15)$)
    ;
    
    \node[circle,fill=BurntOrange,inner sep=1.5pt] (q_null) at (\x6, 0) {};
    \node[circle,fill=black,inner sep=1.5pt] at (\x6, -0.5) {};
    \node[circle,fill=black,inner sep=1.5pt] at (\x6, -1) {};
    \node[circle,fill=black,inner sep=1.5pt] at (\x6, -1.5) {};
    \node[circle,fill=black,inner sep=1.5pt] at (\x6, -2) {};
    \node[circle,fill=black,inner sep=1.5pt] (q_state_all) at (\x6, -2.5) {};
    \node[word] at (\x6, -3.2) {?};
    \draw[thick,dashed,rounded corners,Gray]
        ($(q_null.north west)+(-0.15,0.15)$)
        rectangle
        ($(q_state_all.south east)+(0.15,-0.15)$)
    ;

    \node[anchor=west] at (\x7, 0) {$\emptyset$};
    \node[anchor=west] at (\x7, -0.5) {exclude};
    \node[anchor=west] at (\x7, -1) {next\_to\_2};
    \node[anchor=west] at (\x7, -1.5) {stateid};
    \node[anchor=west] at (\x7, -2) {state\_all};
    \node[anchor=west] at (\x7, -2.5) {area\_1};
    
    \draw[->,thick,dashed,BurntOrange] (root) to (which_null);
    \draw[->,thick,dashed,BurntOrange] (root) to (do_null);
    \draw[->,thick,dashed,BurntOrange] (root) to [bend left=10] (q_null);
    \draw[->,thick,Red] (root) to [bend left=23] (not_exclude);
    \draw[->,thick,Red] (not_exclude) to (states_state_all);
    \draw[->,thick,Red] (not_exclude) to(border_border);
    \draw[->,thick,Red] (border_border) to (texas_stateid);
\end{tikzpicture}%
            \caption{}
            \label{fig:gvcsa}
    \end{subfigure}%
    \caption{%
    \label{fig:reduction}
    \textbf{(a)} Example of a sentence and its associated AST (solid arcs) from the \geoquery{} dataset. The dashed edges indicate predicates and entities anchors (note that this information is not available in the dataset).
    \textbf{(b)} The corresponding generalized valency-constrained not-necessarily-spanning arborescence (red arcs). The root is the isolated top left vertex. Adding $\emptyset$ tags and dotted orange arcs produces a generalized spanning arborescence. 
}
\end{figure*}
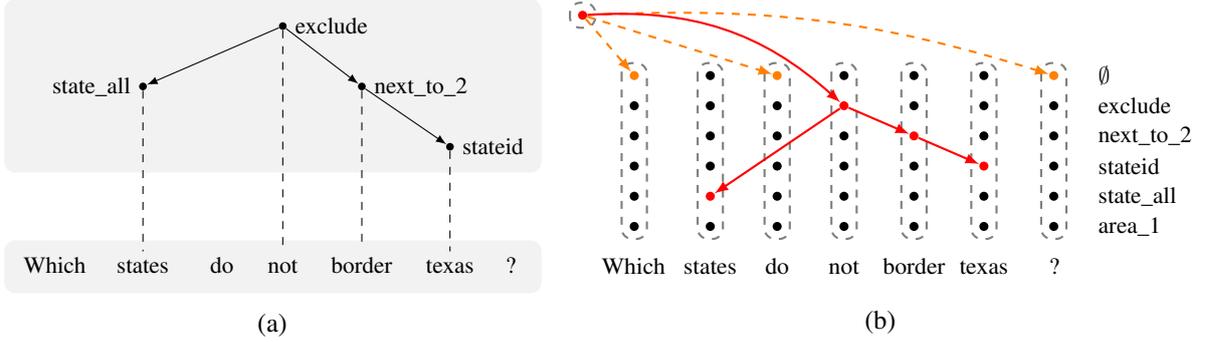
\label{sec:sem_parsing}

We propose to reduce semantic parsing to parsing the abstract syntax tree (AST) associated to a semantic program.
We focus on semantic programs whose ASTs do not have any reentrancy, \emph{i.e.}\ a single predicate or entity cannot be the argument of two different predicates.
Moreover, we assume that each predicate or entity is anchored on exactly one word of the sentence and each word can be the anchor of at most one predicate or entity.
As such, the semantic parsing problem can be reduced to assigning predicates and entities to words and identifying arguments via dependency relations, see Figure~\ref{fig:ast}.
In order to formalize our approach to the semantic parsing problem, we will use concepts from graph theory.
We therefore first introduce the vocabulary and notions that will be useful in the rest of this article.
Notably, the notions of cluster and generalized arborescence will be used to formalize our prediction problem.


\textbf{Notations and definitions.}
Let $\graph = \langle V, A \rangle$ be a directed graph with vertices $V$ and arcs $A \subseteq V \times V$.
An arc in $A$ from a vertex $u \in V$ to a vertex $v \in V$ is denoted either $a \in A$ or $u \to v \in A$.
For any subset of vertices $U \subseteq V$, we denote $\sigma_G^+(U)$ (resp.\ $\sigma_G^-(U)$) the set of arcs leaving one vertex of $U$ and entering one vertex of $V \setminus U$ (resp.\ leaving one vertex of $V \setminus U$ and entering one vertex of $U$) in the graph $\graph$.
Let $B \subseteq A$ be a subset of arcs.
We denote $V[B]$ the cover set of $B$, \emph{i.e.}\ the set of vertices that appear as an extremity of at least one arc in $B$.
A graph $\graph = \langle V, A \rangle$ is an arborescence\footnote{%
In the NLP community, arborescences are often called (directed) trees. We stick with the term arborescence as it is more standard in the graph theory literature, see for example \citet{schrijver2003combinatorial}. Using the term tree introduces a confusion between two unrelated algorithms, Kruskal's maximum spanning tree algorithm \cite{kruskal1956mst} that operates on undirected graphs and Edmond's maximum spanning arborescence algorithm \cite{edmonds1967msa} that operates on directed graphs. Moreover, this prevents any confusion between the graph object called arborescence and the semantic structure called AST.
} rooted at $u \in V$ if and only if (iff) it contains $|V| - 1$ arcs and there is a directed path from $u$ to each vertex in $V$.
In the rest of this work, we will assume that the root is always vertex $0 \in V$.
Let $B \subseteq A$ be a set of arcs such that $\graph' = \langle V[B], B \rangle$ is an arborescence.
Then $G'$ is a spanning arborescence of $\graph$ iff $V[B] = V$.

Let $\pi = \{V_0, ..., V_n\}$ be a partition of $V$ containing $n+1$ clusters.
$\graph'$ is a generalized not-necessarily-spanning arborescence (resp.\ generalized spanning arborescence) on the partition $\pi$ of $\graph$ iff $\graph'$ is an arborescence and $V[B]$ contains at most one vertex per cluster in $\pi$ (resp.\ contains exactly one).

Let $W \subseteq V$ be a set of vertices.
Contracting $W$ consists in replacing in $\graph$ the set $W$ by a new vertex $w \notin V$, replacing all the arcs $u \to v \in \sigma^{-}(W)$ by an arc $u \to w$ and all the arcs $u \to v \in \sigma^{+}(W)$ by an arc $w \to v$.
Given a graph with partition $\pi$, the contracted graph is the graph where each cluster in $\pi$ has been contracted.
While contracting a graph may introduce parallel arcs, it is not an issue in practice, even for weighted graphs.


\subsection{Semantic grammar and AST.}
\label{sec:ast}

The semantic programs we focus on take the form of a functional language, \emph{i.e.}\ a representation where each predicate is a function that takes other predicates or entities as arguments.
The semantic language is typed in the same sense than in ``typed programming languages''.
For example, in \geoquery{}, the predicate \texttt{capital\_2} expects an argument of type \texttt{city} and returns an object of type \texttt{state}.
In the datasets we use, the typing system disambiguates the position of arguments in a function: for a given function, either all arguments are of the same type or the order of arguments is unimportant --- an example of both is the predicate \texttt{intersection\_river} in \geoquery{} that takes two arguments of type \texttt{river}, but the result of the execution is unchanged if the arguments are swapped.\footnote{%
There are a few corner cases like \texttt{exclude\_river}, for which we simply assume arguments are in the same order as they appear in the input sentence.
}

Formally, we define the set of valid semantic programs as the set of programs that can be produced with a semantic grammar $\lang = \langle \tags, \types, \ftype, \fargs \rangle$ where:
\begin{itemize}
    \item $\tags$ is the set of predicates and entities, which we will refer to as the set of tags --- w.l.o.g.\ we assume that $\textsc{Root} \notin E$ 
    where \textsc{Root} is a special tag used for parsing;
    \item $\types$ is the set of types;
    \item $\ftype: \tags \to \types$ is a typing function that assigns a type to each tag;
    \item $\fargs: \tags \times \types \to \N$ is a valency function that assigns the numbers of expected arguments of a given type to each tag.
\end{itemize}
A tag $e \in E$ is an entity iff $\forall t \in T: \fargs(e, t) = 0$.
Otherwise, $e$ is a predicate.

A semantic program in a functional language can be equivalently represented as an AST, a graph where instances of predicates and entities are represented as vertices and where arcs identify arguments of predicates.
Formally, an AST is a labeled graph $\graph = \langle V, A, \flabeling \rangle$ where function $\flabeling: V \to \tags$ assigns a tag to each vertex and arcs identify the arguments of tags, see Figure~\ref{fig:ast}.
An AST $\graph$ is well-formed with respect to the grammar $\lang$ iff $\graph$ is an arborescence and the valency and type constraints are satisfied, \emph{i.e.}\ $\forall u \in V, t \in \types$: 
\[ \fargs(\flabeling(u), t) = | \sigma_G^+(\{u\},t) | \]
where:
\begin{align*}
 \sigma_G^+(\{u\},t)
 = \left\{ 
    \begin{array}{l}
         u \rightarrow v \in \sigma_G^+(\{u\}) \\
         \text{s.t.}~~\ftype(\flabeling(v)) = t
    \end{array}
 \right\}.
\end{align*}

\subsection{Problem reduction and complexity}
\label{sec:reduction}

In our setting, semantic parsing is a joint sentence tagging and dependency parsing problem \cite{bohnet-nivre-2012-transition,li-etal-2011-joint,corro-etal-2017-efficient}:
each content word (\emph{i.e.}\ words that convey a semantic meaning) must be tagged with a predicate or an entity, and dependencies between content words identify arguments of predicates, see Figure~\ref{fig:ast}.
However, our semantic parsing setting differs from standard syntactic analysis in two ways:
(1) the resulting structure is not-necessarily-spanning, there are words (\emph{e.g.}\ function words) that must not be tagged and that do not have any incident dependency --- and those words are not known in advance, they must be identified jointly with the rest of the structure;
(2) the dependency structure is highly constrained by the typing mechanism, that is the predicted structure must be a valid AST.
Nevertheless, similarly to aforementioned works, our parser is graph-based, that is for a given input we build a (complete) directed graph and decoding is reduced to computing a constrained subgraph of maximum weight.

Given a sentence $\vw = \evw_1 ... \evw_n$ with $n$ words and a grammar $\lang$, we construct a clustered labeled graph $\graph = \langle V, A, \pi, \felabeling \rangle$ as follows.
The partition $\pi = \{V_0, ..., V_n\}$ contains $n+1$ clusters, where $V_0$ is a root cluster and each cluster $V_i$, $i \neq 0$, is associated to word $\evw_i$.
The root cluster $V_0 = \{ 0 \}$ contains a single vertex that will be used as the root and every other cluster contains $|E|$ vertices.
The extended labeling function $\felabeling: V \rightarrow E \cup \{ \textsc{Root} \}$ assigns a tag in $E$ to each vertex $v \in V \setminus \{ 0 \}$ and \textsc{Root} to vertex $0$. Distinct vertices in a cluster $V_i$ cannot have the same label, \emph{i.e.}\ $\forall u, v \in V_i: u \neq v \implies \felabeling(u) \neq \felabeling(v)$.

Let $B \subseteq A$ be a subset of arcs.
The graph $\graph' = \langle V[B], B \rangle$ defines a 0-rooted generalized valency-constrained not-necessarily-spanning arborescence iff it is a generalized arborescence of $\graph$, there is exactly one arc leaving $0$ and the sub-arborescence rooted at the destination of that arc is a valid AST with respect to the grammar $\lang$.
As such, there is a one-to-one correspondence between ASTs anchored on the sentence $\vw$ and generalized valency-constrained not-necessarily-spanning arborescences in the graph $\graph$, see Figure~\ref{fig:gvcsa}.

For any sentence $\vw$, our aim is to find the AST that most likely corresponds to it.
Thus, after building the graph $\graph$ as explained above, the neural network described in Appendix~\ref{app:exp} is used to produce a vector of weights $\vmu \in \mathbb{R}^{|V|}$ associated to the set of vertices $V$ and a vector of weights $\vphi \in \mathbb{R}^{|A|}$ associated to the set of arcs $A$.
Given these weights, graph-based semantic parsing is reduced to an optimization problem called the maximum generalized valency-constrained not-necessarily-spanning arborescence (MGVCNNSA) in the graph $\graph$.


\begin{theorem}
\label{th:parsing}
The MGVCNNSA problem is NP-hard.
\end{theorem}
\noindent The proof is in Appendix~\ref{proofs}.

\subsection{Mathematical program}
\label{sec:ilp_definition}

Our graph-based approach to semantic parsing has allowed us to prove the intrinsic hardness of the problem.
We follow previous work on graph-based parsing \cite{martins-etal-2009-concise,koo-etal-2010-dual}, \emph{inter alia}, by proposing an integer linear programming (ILP) formulation in order to compute (approximate) solutions.

Remember that in the joint tagging and dependency parsing interpretation of the semantic parsing problem, the resulting structure is not-necessarily-spanning, meaning that some words may not be tagged.
In order to rely on well-known algorithms for computing spanning arborescences as a subroutine of our approximate solver, 
we first introduce the notion of extended graph.
Given a graph $G = \langle V, A, \pi, \felabeling \rangle$, we construct an extended graph $\overline G = \langle \overline V, \overline A, \overline \pi, \felabeling \rangle$\footnote{The labeling function is unchanged as there is no need for types for vertices in $\overline{V} \setminus V$.} containing $n$ additional vertices $\{ \overline{1}, ..., \overline{n} \}$ that are distributed along clusters, \emph{i.e.}\ $\overline{\pi} = \{ V_0, V_1 \cup \{ \overline{1} \}, ..., V_n \cup \{ \overline{n} \} \}$, and arcs from the root to these extra vertices, \emph{i.e.}\ $\overline{A} = A \cup \{ 0 \to \overline{i} | 1 \leq i \leq n\}$.
Let $B \subseteq A$ be a subset of arcs such that $\langle V[B], B \rangle$ is a generalized not-necessarily-spanning arborescence on $G$.
Let $\overline B \subseteq \overline A$ be a subset of arcs defined as $\overline B = B \cup \{ 0 \to \overline{i} | \sigma^-_{\langle V[B], B \rangle}(V_i) = \emptyset \}$.
Then, there is a one-to-one correspondence between generalized not-necessarily-spanning arborescences $\langle V[B], B \rangle$ and generalized spanning arborescences $\langle \overline V[\overline B], \overline B \rangle$, see Figure~\ref{fig:gvcsa}.

Let $\vx \in \{0, 1\}^{|\overline V|}$ and $\vy \in \{0, 1\}^{|\overline A|}$ be variable vectors indexed by vertices and arcs such that a vertex $v \in V$ (resp.\ an arc $a \in A$) is selected iff $\evx_v = 1$ (resp.\ $\evy_a = 1$).
The set of 0-rooted generalized valency-constrained spanning arborescences on $\overline{G}$ can be written as the set of variables $\langle \vx, \vy \rangle$ satisfying the following linear constraints.
First, we restrict $\vy$ to structures that are spanning arborescences over $\overline G$ where clusters have been contracted:
\begin{align}
&\sum_{a \in \sigma_{\overline G }^-(V_0)} \evy_a = 0 
    && \label{eq:sa_first} \\
&\sum_{a \in \sigma_{\overline G }^-\left(
    \bigcup_{\overline{U} \in \overline{\pi'}}
    \overline U
    \right)} \evy_a \geq 1
    &&  \forall \overline{\pi'} \subseteq \overline{\pi} \setminus \{ V_0 \}
    \label{eq:sa_connectivity}
\end{align}
\begin{align}
&\sum_{a \in \sigma_{\overline G }^-(\overline{V_i})} \evy_a = 1 
    && \forall \overline{V_i} \in \overline \pi \setminus \{ V_0 \} \label{eq:sa_singlehead}
\end{align}
Constraints~(\ref{eq:sa_connectivity}) ensure that the contracted graph is weakly connected.
Constraints~(\ref{eq:sa_singlehead}) force each cluster to have exactly one incoming arc.
The set of vectors $\vy$ that satisfy these three constraints are exactly the set of $0$-rooted spanning arborescences on the contracted graph, see~\citet[][Section 52.4]{schrijver2003combinatorial} for an in-depth analysis of this polytope.
The root vertex is always selected and other vertices are selected iff they have one incoming selected arc:
\begin{align}
&\evx_0 = 1 
    &&  \label{eq:root_vertex} \\
&\evx_u = \sum_{a \in \sigma_{\overline G }^-(\{u\})} \evy_a 
    && \hspace{-0.5cm} \forall u \in \overline{V} \setminus \{0\} \label{eq:vertex}
\shortintertext{Note that constraints (\ref{eq:sa_first})--(\ref{eq:sa_singlehead}) do not force selected arcs to leave from a selected vertex as they operate at the cluster level.
This property will be enforced via the valency constraints:}
&\sum_{u \in V \setminus \{0\}} \evy_{0 \to u} = 1 && & \label{eq:predicate} \\
&\sum_{a \in \sigma^+_G(\{u\}, t)} \hspace{-0.5cm}\evy_{a}= \evx_u \fargs(l(u), t) 
    && \begin{array}[t]{@{}l@{}l@{}}
         \forall& t \in T, \\
                & u \in V \setminus \{0\}
      \end{array} \hspace{-2cm} \label{eq:valency}
\end{align}
Constraint (\ref{eq:predicate}) forces the root to have exactly one outgoing arc into a vertex $u \in V \setminus \{0\}$ (\emph{i.e.}\ a vertex that is not part of the extra vertices introduced in the extended graph) that will be the root of the AST.
Constraints (\ref{eq:valency}) force the selected vertices and arcs to produce a well-formed AST with respect to the grammar $\lang$.
Note that these constraints are only defined for vertices in $V \setminus \{0\}$, \emph{i.e.}\ they are neither defined for the root vertex nor for the extra vertices introduced in the extended graph.

To simplify notations, we introduce the following sets:
\begin{align*}
    \spanset
    &= \left\{\begin{array}{l}
    \langle \vx, \vy \rangle \in \{0, 1\}^{\overline{V}} \times \{0, 1\}^{\overline{A}} \\
    \text{s.t.~} \vx\text{~and~}\vy\text{~satisfy~(\ref{eq:sa_first})--(\ref{eq:vertex})}
    \end{array}\right\},
    \\
    \valset
    &= \left\{\begin{array}{l}
    \langle \vx, \vy \rangle \in \{0, 1\}^{\overline{V}} \times \{0, 1\}^{\overline{A}} \\
    \text{s.t.~} \vx\text{~and~}\vy\text{~satisfy~(\ref{eq:predicate})--(\ref{eq:valency})}
    \end{array}\right\},
\end{align*}
and $\mathcal C = \spanset \cap \valset$.
Given vertex weights $\vmu \in \R^{|\overline{V}|}$ and arc weights $\vphi \in \R^{|\overline{A}|}$, computing the MGVCNNSA is equivalent to solving the following ILP:
\begin{align*}
    \textsc{(Ilp1)} \quad \max_{\vx, \vy}\quad& \vmu^\top \vx + \vphi^\top \vy
    \\
    \text{s.t.}\quad& \langle \vx, \vy \rangle \in \spanset \text{~and~} \langle \vx, \vy \rangle \in \valset
\end{align*}
Without constraint $\langle \vx, \vy \rangle \in \valset$, the problem would be easy to solve.
The set $\spanset$ is the set of spanning arborescences over the contracted graph, hence to maximize over this set we can simply:
(1) contract the graph and assign to each arc in the contracted graph the weight of its corresponding arc plus the weight of its destination vertex in the original graph;
(2) run the the maximum spanning arborescence algorithm \cite[MSA,][]{edmonds1967msa,tarjan1977msa} on the contracted graph, which has a $\mathcal O(n^2)$ time-complexity.
This process is illustrated on Figure~\ref{fig:pipeline} (top).
Note that the contracted graph may have parallel arcs, which is not an issue in practice as only the one of maximum weight can appear in a solution of the MSA.

We have established that MAP inference in our semantic parsing framework is a NP-hard problem.
We proposed an ILP formulation of the problem that would be easy to solve if some constraints were removed.
This property suggests the use of an approximation algorithm that introduces the difficult constraints as penalties.
As a similar setting arises from our weakly supervised loss function,
the presentation of the approximation algorithm is deferred until Section~\ref{sec:opt}.
\section{Training objective functions}
\label{sec:training}

\subsection{Supervised training objective}
\label{sec:suploss}

We define the likelihood of a pair $\langle \vx, \vy \rangle \in \fullset$ via the Boltzmann distribution:
\begin{align*} 
p_{\vmu, \vphi}(\vx, \vy) 
 &= \exp(\vmu^\top\vx + \vphi^\top\vy - c(\vmu, \vphi)), \\
\intertext{where $c(\vmu, \vphi)$ is the log-partition function:}
c(\vmu, \vphi)
&= \log \sum_{\langle \vx', \vy' \rangle \in \fullset} \exp(\vmu^\top\vx' + \vphi^\top \vy').
\end{align*}
During training, we aim to maximize the log-likelihood of the training dataset.
The log-likelihood of an observation $\langle \vx, \vy\rangle$ is defined as:
\begin{align*} 
    \suploss(\vmu, \vphi; \vx, \vy)
    &= \log p_{\vmu, \vphi}(\vx, \vy) \\
    &= \vmu^\top\vx + \vphi^\top\vy - c(\vmu, \vphi).
\end{align*}
Unfortunately, computing the log-partition function is intractable as it requires summing over all feasible solutions.
Instead, we rely on a surrogate lower-bound as an objective function.
To this end, we derive an upper bound (because it is negated in $\suploss$) to the second term:
{\color{black}
a sum of log-sum-exp functions that sums over each cluster of vertices independently and over incoming arcs in each cluster independently, which is tractable.
This loss can be understood as a generalization of the head selection loss used in dependency parsing \cite{zhang2017head}.
We now detail the derivation and prove that it is an upper bound to the log-partition function.
}

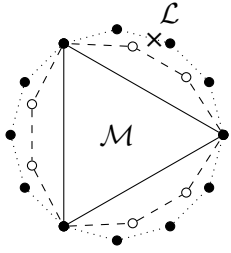
\begin{figure}
    \centering
    \begin{tikzpicture}[
    scale=0.7,
    every node/.style={
        circle,
        inner sep=0pt,
        minimum size=0.12cm,
        draw=black,
    },
    integral_node/.style={
        fill=black
    },
    non_integral_node/.style={
        fill=white
    }
]
\node[integral_node] (p3_0) at (2.000000, 0.000000) {};
\node[integral_node] (p3_1) at (-1.000000, 1.732051) {};
\node[integral_node] (p3_2) at (-1.000000, -1.732051) {};
\draw[solid] (p3_0) -- (p3_1);
\draw[solid] (p3_1) -- (p3_2);
\draw[solid] (p3_2) -- (p3_0);
\node[draw=none] at (0, 0) {$\mathcal M$};

\node[integral_node] (p9_0) at (2.000000, 0.000000) {};
\node[non_integral_node] (p9_1) at (1.302276, 1.092739) {};
\node[non_integral_node] (p9_2) at (0.295202, 1.674173) {};
\node[integral_node] (p9_3) at (-1.000000, 1.732051) {};
\node[non_integral_node] (p9_4) at (-1.597477, 0.581434) {};
\node[non_integral_node] (p9_5) at (-1.597477, -0.581434) {};
\node[integral_node] (p9_6) at (-1.000000, -1.732051) {};
\node[non_integral_node] (p9_7) at (0.295202, -1.674173) {};
\node[non_integral_node] (p9_8) at (1.302276, -1.092739) {};
\draw[dashed] (p9_0) -- (p9_1);
\draw[dashed] (p9_1) -- (p9_2);
\draw[dashed] (p9_2) -- (p9_3);
\draw[dashed] (p9_3) -- (p9_4);
\draw[dashed] (p9_4) -- (p9_5);
\draw[dashed] (p9_5) -- (p9_6);
\draw[dashed] (p9_6) -- (p9_7);
\draw[dashed] (p9_7) -- (p9_8);
\draw[dashed] (p9_8) -- (p9_0);

\node[integral_node] (p12_0) at (2.000000, 0.000000) {};
\node[integral_node] (p12_1) at (1.732051, 1.000000) {};
\node[integral_node] (p12_2) at (1.000000, 1.732051) {};
\node[integral_node] (p12_3) at (0.000000, 2.000000) {};
\node[integral_node] (p12_4) at (-1.000000, 1.732051) {};
\node[integral_node] (p12_5) at (-1.732051, 1.000000) {};
\node[integral_node] (p12_6) at (-2.000000, 0.000000) {};
\node[integral_node] (p12_7) at (-1.732051, -1.000000) {};
\node[integral_node] (p12_8) at (-1.000000, -1.732051) {};
\node[integral_node] (p12_9) at (-0.000000, -2.000000) {};
\node[integral_node] (p12_10) at (1.000000, -1.732051) {};
\node[integral_node] (p12_11) at (1.732051, -1.000000) {};
\draw[dotted] (p12_0) -- (p12_1);
\draw[dotted] (p12_1) -- (p12_2);
\draw[dotted] (p12_2) -- node[near start,draw=none,label={above,xshift=0.2cm:$\mathcal L$}] {\texttimes} (p12_3);
\draw[dotted] (p12_3) -- (p12_4);
\draw[dotted] (p12_4) -- (p12_5);
\draw[dotted] (p12_5) -- (p12_6);
\draw[dotted] (p12_6) -- (p12_7);
\draw[dotted] (p12_7) -- (p12_8);
\draw[dotted] (p12_8) -- (p12_9);
\draw[dotted] (p12_9) -- (p12_10);
\draw[dotted] (p12_10) -- (p12_11);
\draw[dotted] (p12_11) -- (p12_0);

\end{tikzpicture}
    
    \caption{%
    {\color{black}Polyhedrons illustration.
    The solid lines represent the convex hull of feasible solutions of $\textsc{Ilp1}$, denoted $\mathcal M$, whose vertices are feasible integer solutions (black vertices).
    The dashed lines represent the convex hull of feasible solutions of the linear relaxation of $\textsc{Ilp1}$, which has non integral vertices (in white).
    Finally, the dotted lines represent the polyhedron $\mathcal L$ that is used to approximate $c(\vmu, \vphi)$.
    All its vertices are integral, but some of them are not feasible solutions of $\textsc{Ilp1}$.
    }}
    \label{fig:polyhedron}
\end{figure}

{\color{black}
Let $\mU$ be a matrix such that each row contains a pair $\langle \vx, \vy \rangle \in \fullset$ and $\Delta^{|\fullset|}$ be the simplex of dimension $|\fullset|-1$, \emph{i.e.} the set of all stochastic vectors of dimension $|\fullset|$.
}
The log-partition function can then be rewritten {\color{black}using its variational formulation}:
\begin{align*}
c(\vmu, \vphi) = \max_{\vp \in \Delta^{|\fullset|}} \vp^\top \left( \mU \begin{bmatrix}\vmu \\ \vphi\end{bmatrix} \right) + H[\vp],
\end{align*}
where $H[\vp] = -\sum_i \evp_i \log \evp_i$ is the Shannon entropy.
We refer the reader to \citet[][Example 3.25]{boyd2004convex}, \citet[][Section 3.6]{wainwright2008exp} and \citet[][Section 4.4.10]{beck2017first}.
{\color{black}Note that this formulation remains impractical as $\vp$ has an exponential size.}
Let $\marginal = \conv(\fullset)$ be the marginal polytope, \emph{i.e.}\ the convex hull of the feasible integer solutions, we can rewrite the above variational formulation as:
\begin{align*}
& = \max_{\vm \in \marginal} \vm^\top \begin{bmatrix}\vmu \\ \vphi\end{bmatrix} + H_{\marginal}[\vm]
\end{align*}
where $H_{\marginal}$ is a joint entropy function defined such that the equality holds.
{\color{black}
The maximization in this reformulation acts on the marginal probabilities of parts (vertices and arcs) and has therefore a polynomial number of variables.
}
We refer the reader to \citet[][5.2.1]{wainwright2008exp} and \citet[][Section 7]{blondel2020fy} for more details.
Unfortunately, this optimization problem is hard to solve as $\marginal$ cannot be caracterized in an explicit manner and $H_{\marginal}$ is defined indirectly and lacks a polynomial closed form \cite[Section 3.7]{wainwright2008exp}.
However, we can derive an upper bound to the log-partition function by
decomposing the entropy term $H_{\marginal}$ \cite[][Property 4 on page 41, \emph{i.e.}\ $H$ is an upper bound]{cover1999infoth}
and
by using an outer approximation to the marginal polytope $\relmarginal \supseteq \marginal$ (\emph{i.e.}\ increasing the search space):
\begin{align*}
& \leq \max_{\vm \in \relmarginal} \vm^\top \begin{bmatrix}\vmu \\ \vphi\end{bmatrix} + H[\vm]
\end{align*}
In particular, we observe that each pair $\langle \vx, \vy \rangle \in \fullset $ has exactly one vertex selected per cluster $\overline{V_i} \in \overline \pi$ and one incoming arc selected per cluster $\overline{V_i} \in \overline \pi \setminus \{V_0\}$.
We denote $\oneset$ the set of all the pairs $\langle \vx, \vy \rangle$ that satisfy these constraints.
By using $\relmarginal = \conv(\oneset)$ as an outer approximation to the marginal polytope (see Figure~\ref{fig:polyhedron}) the optimization problem can be rewritten as a sum of independent problems.
As each of these problems is the variational formulation of a log-sum-exp term, the upper bound on $c(\vmu, \vphi)$ can be expressed as a sum of log-sum-exp functions, one over vertices in each cluster $\overline{V_i} \in \overline \pi  \setminus \{V_0\}$ and one over incoming arcs $\sigma_{\overline{G}}^-(\overline{V_i})$ for each cluster $\overline{V_i} \in \overline \pi \setminus \{V_0\}$.
Although this type of approximation may not result in a Bayes consistent loss \cite{corro2023eacl}, it works well in practice.

\subsection{Weakly-supervised training objective}
\label{sec:weaklloss}

Unfortunately, training data often does not include gold pairs $\langle \vx, \vy \rangle$ but instead only the AST, without word anchors (or word alignment).
This is the case for the three datasets we use in our experiments.
We thus consider our training signal to be the set of all structures that induce the annotated AST, which we denote $\goldset$.

The weakly-supervised loss is defined as:
\begin{align*}
\weakloss(\vmu, \vphi; \goldset)
&= \log \sum_{\langle \vx, \vy \rangle \in \goldset} p_{\vmu, \vphi}(\vx, \vy),
\end{align*}
\emph{i.e.}\ we marginalize over all the structures that induce the gold AST. We can rewrite this loss as:
\begin{align*}
= \left( \log \sum_{\langle \vx, \vy \rangle \in \goldset} \exp(\vmu^\top\vx + \vphi^\top\vy) \right) - c(\vmu, \vphi).
\end{align*}
The two terms are intractable.
We approximate the second term using the bound defined in Section~\ref{sec:suploss}.

We now derive a tractable lower bound to the first term.
Let $q$ be a proposal distribution such that $q(\vx, \vy) = 0$ if $\langle \vx, \vy \rangle \notin \goldset$.
We derive the following lower bound via Jensen's inequality:
\begin{align*}
\log \sum_{\langle \vx, \vy \rangle \in \goldset} \exp(\vmu^\top\vx + \vphi^\top \vy)&\\
&\hspace{-4cm}= \log \E_{q} \left[ \frac{\exp(\vmu^\top\vx + \vphi^\top \vy))}{q(\vx, \vy)} \right] \\
&\hspace{-4cm} \geq \E_{q} \left[\vmu^\top\vx + \vphi^\top \vy\right] + H[q].
\end{align*}
This bound holds for any distribution $q$ satisfying the aforementioned condition.
We choose to maximize this lower bound using a distribution that gives a probability of one to a single structure, as in ``hard'' EM \cite[][Section 6]{neal1998em}.

For a given sentence $\vw$, let $G = \langle V, A, \pi, \felabeling \rangle$ be a graph defined as in Section~\ref{sec:reduction} and $G' = \langle V', A', \flabeling'\rangle$ be an AST defined as in Section~\ref{sec:ast}.
We aim to find the GVCNNSA in $G$ of maximum weight whose induced AST is exactly $G'$.
This is equivalent to aligning each vertex in $V'$ with one vertex of $V \setminus \{0\}$ s.t.\ there is at most one vertex per cluster of $\pi$ appearing in the alignment and where the weight of an alignment is defined as:
\begin{enumerate}
    \item for each vertex $u' \in V'$, we add the weight of the vertex $u \in V$ it is aligned to --- moreover, if $u'$ is the root of the AST we also add the weight of the arc $0 \to u$;
    \item for each arc $u' \to v' \in A'$, we add the weight of the arc $u \to v$ where $u \in V$ (resp.\ $v \in V$) is the vertex $u'$ (resp. $v'$) it is aligned with.
\end{enumerate}
Note that this {\color{black}latent anchoring inference} consists in computing a (partial) alignment between vertices of $G$ and $G'$, but the fact that we need to take into account arc weights forbids the use of the Kuhn–Munkres algorithm \cite{Kuhn1955Hungarian}.

\begin{algorithm*}[t]
\caption{\label{alg:assignment}%
Unconstrained alignment of maximum weight between a graph $G$ and an AST $G'$
}
\small
\begin{algorithmic}
\Function{DPAlignment}{$G, G'$}
     \For{$u' \in V'$\text{~in reverse topological order}}
        \For{$u \in \{ v \in V | \felabeling(v) = \flabeling'(u') \}$}
        \Comment{We can only map $u'$ to vertices $u$ if they have the same tag.}
            \State $
                \textsc{Chart}[u', u] \gets
                \mu_{u} + 
                \sum_{u' \to v' \in \sigma_{G'}^+(\{u'\})}
                    \big(
                    \max_{v \in V}
                    \textsc{Chart}[v', v] + \phi_{u \to v}
                    \big)
            $
        \EndFor
     \EndFor
     \Return $\max_{u \in V} \textsc{Chart}[r', u] + \phi_{0 \to u}$
     \Comment{Where $r' \in A'$ is the root of the AST.}
\EndFunction
\end{algorithmic}
\end{algorithm*}

\begin{algorithm*}[t]
\caption{\label{alg:fw}%
Conditional gradient
}
\small
\begin{algorithmic}
\Function{ConditionalGradient}{$G, G'$}
    \State Let $\vz^{(0)} \in \conv(\mathcal{C}^{\text{(easy)}})$
     \For{$k \in \{ 0 .... K\}$}
        \Comment{Where $K$ is the maximum number of iterations}
        \State $\vd \gets \big(\lmo_{\conv(\mathcal C^{\text{(easy)}})} (\nabla         g(\vz^{(k)})) \big) - \vz^{(k)}$
        \Comment{Compute the update direction}
        \If{$\nabla g(\vz^{(k)})^\top \vd \leq \epsilon$}
            \Return $\vz^{(k)}$
            \Comment{If the dual gap is small, $\vz^{(k)}$ is (almost) optimal}
        \EndIf
        \State $\gamma \in \argmax_{\gamma \in [0, 1]} g(\vz^{(k)} + \gamma \vd)$
        \Comment{Compute or approximate the optimal stepsize}
        \State $\vz^{(k + 1)} = \vz^{(k)} + \gamma \vd$
        \Comment{Update the current point}
     \EndFor
    \Return $\vz^{(k)}$
\EndFunction
\end{algorithmic}
\end{algorithm*}

\begin{theorem}
    \label{th:assignment}
    Computing the anchoring of maximum weight of an AST with a graph $G$ is NP-hard.
\end{theorem}
\noindent The proof is in Appendix~\ref{proofs}.

Therefore, we propose an optimization-based approach to compute the distribution $q$.
Note that the problem has a constraint requiring each cluster $V_i \in \pi$ to be aligned with at most one vertex $v' \in V'$, \emph{i.e.}\ each word in the sentence can be aligned with at most one vertex in the AST.
If we remove this constraint, then the problem becomes tractable via dynamic programming.
Indeed, we can recursively construct a table $\textsc{Chart}[u', u]$, $u' \in V'$ and $u \in V$, containing the score of aligning vertex $u'$ to vertex $u$ plus the score of the best alignment of all the descendants of $u'$.
To this end, we simply visit the vertices $V'$ of the AST in reverse topological order, see Algorithm \ref{alg:assignment}.
The best alignment can be retrieved via back-pointers.

Computing $q$ is therefore equivalent to solving the following ILP:
\begin{align*}
    \textsc{(Ilp2)}\quad\max_{\vx, \vy} \quad& \vmu^\top \vx + \vphi^\top \vy \\
    \text{s.t.} \quad & \langle \vx, \vy \rangle \in \mathcal{C}^{*\text{(relaxed)}}, \\
    & \sum_{u \in V_i} \evx_u \leq 1 \qquad \forall V_i \in \pi.
\end{align*}
The set $\mathcal{C}^{\text{*(relaxed)}}$ is the set of feasible solutions of the dynamic program in Algorithm~\ref{alg:assignment}, whose convex hull can be described via linear constraints \cite{martin1990polyhedral}.
\section{Efficient inference}
\label{sec:opt}

\begin{figure*}
    \centering
%

\definecolor{Red}{rgb}{1, 0, 0}
\definecolor{Gray}{rgb}{0.5, 0.5, 0.5}
\definecolor{BurntOrange}{rgb}{1, 0.5, 0}

\tikzstyle{word} = [rectangle, rounded corners, text width=5.5em, text centered, minimum height=2em, anchor=mid]

\begin{tikzpicture}[
    score/.style={midway,inner sep=0pt,fill=white,circle},
    scale=0.9,every node/.style={scale=0.9}
]
\node[circle,fill=black,inner sep=1.5pt] (root) at (-0.85, 1) {};

\draw[thick,dashed,rounded corners,Gray]
    ($(root.north west)+(-0.15,0.15)$)
    rectangle
    ($(root.south east)+(0.15,-0.15)$)
;

\node[anchor=east] at (-1, 0) {$\emptyset$};
\node[anchor=east] at (-1, -1) {state\_all};
\node[anchor=east] at (-1, -3) {loc\_1};

%
%
\node[circle,fill=black,inner sep=1.5pt,label={below:\scriptsize $0$}] (list0) at (0, 0) {};
\node[circle,fill=black,inner sep=1.5pt,label={below,yshift=-3pt:\scriptsize $-1$}] (list1) at (0, -1) {};
\node[circle,fill=black,inner sep=1.5pt,label={[fill=white,inner sep=1pt,yshift=-3pt]below:\scriptsize $0{\color{red}-3}$}] (list2) at (0, -3) {};
\node[word] at (0, -4.2) {List};
\begin{pgfonlayer}{bg}
\draw[thick,dashed,rounded corners,Gray]
    ($(list0.north west)+(-0.15,0.15)$)
    rectangle
    ($(list2.south east)+(0.15,-0.55)$)
;
\end{pgfonlayer}
%
%
\node[circle,fill=black,inner sep=1.5pt,label={below:\scriptsize $0$}] (states0) at (3, 0) {};
\node[circle,fill=black,inner sep=1.5pt,label={below:\scriptsize $1$}] (states1) at (3, -1) {};
\node[circle,fill=black,inner sep=1.5pt,label={below:\scriptsize $-1$}] (states2) at (3, -3) {};
\node[word] at (3, -4.2) {states};
\draw[thick,dashed,rounded corners,Gray]
    ($(states0.north west)+(-0.15,0.15)$)
    rectangle
    ($(states2.south east)+(0.15,-0.55)$)
;

\draw[->, dotted] (root) to node[score]{\scriptsize $0$} (list0);
\draw[->, dotted] (root) to[bend right=20] node[score,near end]{\scriptsize $1$} (list1);
\draw[->, dotted] (root) to[bend right=20] node[score,near end]{\scriptsize $1$} (list2);

\draw[->, dotted] (root) to node[score]{\scriptsize $0$} (states0);
\draw[->, dotted] (root) to node[score]{\scriptsize $1.5$} (states1);
\draw[->, dotted] plot [smooth,tension=1] coordinates {(root) (3.5, 0.5) (states2)};
\node at (3.5, 0.5) {\scriptsize $1.5$};

\draw[->, dotted] (list1) to[bend left=20] node[score]{\scriptsize $-1$} (states1);
\draw[->, dotted] (states1) to[bend left=0] node[score]{\scriptsize $-1$} (list1);

\draw[->, dotted] (list2) to[bend left=0] node[score]{\scriptsize $-1$} (states2);
\draw[->, dotted] (states2) to[bend left=20] node[score]{\scriptsize $-1$} (list2);

\draw[->, dotted] (list1) to[bend left=20] node[score,xshift=6pt,yshift=-2pt]{\scriptsize $-1$} (states2);
\draw[->, dotted] (states2) to[bend left=20] node[score,xshift=-6pt,yshift=2pt]{\scriptsize $-1$} (list1);

\draw[->, dotted] (list2) to[bend left=20] node[score,xshift=-6pt,yshift=-2pt,rectangle]{\scriptsize $-1{\color{red}+3}$} (states1);
\draw[->, dotted] (states1) to[bend left=20] node[score,xshift=6pt,yshift=2pt]{\scriptsize $-1$} (list2);

%
%
%
%

\node[circle,fill=black,inner sep=1.5pt] (c1_root) at (4.65, 1) {};
\node[circle,fill=black,inner sep=3pt] (c1_list) at (5.5, 0) {};
\node[word] at (5.5, -0.7) {List};
\node[circle,fill=black,inner sep=3pt] (c1_states) at (7.5, 0) {};
\node[word] at (7.5, -0.7) {states};

\draw[->, solid, thick] (c1_root) to node[score]{\scriptsize $1$} (c1_list);
\draw[->, solid, thick] (c1_root) to[bend left=30] node[score]{\scriptsize $2.5$} (c1_states.north);
\draw[->, dotted] (c1_list) to[bend left=20] node[score]{\scriptsize $0$} (c1_states);
\draw[->, dotted] (c1_states) to[bend left=20] node[score]{\scriptsize $-1$} (c1_list);

%
%
%
%

\node[circle,fill=black,inner sep=1.5pt] (d1_root) at (9.15, 1) {};

\draw[thick,dashed,rounded corners,Gray]
    ($(d1_root.north west)+(-0.15,0.15)$)
    rectangle
    ($(d1_root.south east)+(0.15,-0.15)$)
;

%
%
\node[circle,fill=black,inner sep=1.5pt] (d1_list0) at (10, 0.5) {};
\node[circle,fill=black,inner sep=1.5pt] (d1_list1) at (10, 0) {};
\node[circle,fill=black,inner sep=1.5pt] (d1_list2) at (10, -0.5) {};
\node[word] at (10, -1.1) {List};
\draw[thick,dashed,rounded corners,Gray]
    ($(d1_list0.north west)+(-0.15,0.15)$)
    rectangle
    ($(d1_list2.south east)+(0.15,-0.15)$)
;
%
%

\node[anchor=west] at (11.3, 0.5) {$\emptyset$};
\node[anchor=west] at (11.3, 0) {state\_all};
\node[anchor=west] at (11.3, -0.5) {loc\_1};

\node[circle,fill=black,inner sep=1.5pt] (d1_states0) at (11, 0.5) {};
\node[circle,fill=black,inner sep=1.5pt] (d1_states1) at (11, 0) {};
\node[circle,fill=black,inner sep=1.5pt] (d1_states2) at (11, -0.5) {};
\node[word] at (11, -1.1) {states};
\draw[thick,dashed,rounded corners,Gray]
    ($(d1_states0.north west)+(-0.15,0.15)$)
    rectangle
    ($(d1_states2.south east)+(0.15,-0.15)$)
;

\draw[->,solid,thick] (d1_root) to (d1_list2);
\draw[->,solid,thick] (d1_root) to[bend left=30] (d1_states1);

%
%
%
%

\node[circle,fill=black,inner sep=1.5pt] (c2_root) at (4.65, -2.5) {};
\node[circle,fill=black,inner sep=3pt] (c2_list) at (5.5, -3.5) {};
\node[word] at (5.5, -4.2) {List};
\node[circle,fill=black,inner sep=3pt] (c2_states) at (7.5, -3.5) {};
\node[word] at (7.5, -4.2) {states};

\draw[->, solid, thick] (c2_root) to node[score]{\scriptsize $0$} (c2_list);
\draw[->, solid, thick] (c2_root) to[bend left=30] node[score]{\scriptsize $2.5$} (c2_states.north);
\draw[->, dotted] (c2_list) to[bend left=20] node[score]{\scriptsize $2$} (c2_states);
\draw[->, dotted] (c2_states) to[bend left=20] node[score]{\scriptsize $-1$} (c2_list);

%
%
%
%

\node[circle,fill=black,inner sep=1.5pt] (d2_root) at (9.15, -2.5) {};

\draw[thick,dashed,rounded corners,Gray]
    ($(d2_root.north west)+(-0.15,0.15)$)
    rectangle
    ($(d2_root.south east)+(0.15,-0.15)$)
;

%
%
\node[circle,fill=black,inner sep=1.5pt] (d2_list0) at (10, -3) {};
\node[circle,fill=black,inner sep=1.5pt] (d2_list1) at (10, -3.5) {};
\node[circle,fill=black,inner sep=1.5pt] (d2_list2) at (10, -4) {};
\node[word] at (10, -4.6) {List};
\draw[thick,dashed,rounded corners,Gray]
    ($(d2_list0.north west)+(-0.15,0.15)$)
    rectangle
    ($(d2_list2.south east)+(0.15,-0.15)$)
;
%
%

\node[anchor=west] at (11.3, -3) {$\emptyset$};
\node[anchor=west] at (11.3, -3.5) {state\_all};
\node[anchor=west] at (11.3, -4) {loc\_1};

\node[circle,fill=black,inner sep=1.5pt] (d2_states0) at (11, -3) {};
\node[circle,fill=black,inner sep=1.5pt] (d2_states1) at (11, -3.5) {};
\node[circle,fill=black,inner sep=1.5pt] (d2_states2) at (11, -4) {};
\node[word] at (11, -4.6) {states};
\draw[thick,dashed,rounded corners,Gray]
    ($(d2_states0.north west)+(-0.15,0.15)$)
    rectangle
    ($(d2_states2.south east)+(0.15,-0.15)$)
;

\draw[->,solid,thick] (d2_root) to (d2_list0);
\draw[->,solid,thick] (d2_root) to[bend left=30] (d2_states1);

\draw[-{Triangle[width=8pt,length=6pt]}, line width=4pt] (2,1.4) to[bend left=20] node[midway,yshift=0.4cm]{contract} (5, 1.4);
\draw[-{Triangle[width=8pt,length=6pt]}, line width=4pt] (6.5,1.4) to[bend left=20] node[midway,yshift=0.4cm]{expand} (9.5, 1.4);

\draw[-{Triangle[width=8pt,length=6pt]}, line width=4pt] (2,-4.5) to[bend right=20] node[midway,yshift=-0.4cm]{add penalties + contract} (5, -4.5);
\draw[-{Triangle[width=8pt,length=6pt]}, line width=4pt] (6.5,-4.5) to[bend right=20] node[midway,yshift=-0.4cm]{expand} (9.5, -4.5);

\end{tikzpicture}
    \caption{%
    {\color{black}Illustration of the approximate inference algorithm on the two-word sentence ``List states'', where we assume the grammar has one entity \texttt{state\_all} and one predicate \texttt{loc\_1} that takes exactly one entity as argument.
    The left graph is the extended graph for the sentence, including vertices and arcs weights (in black).
    If we ignore constraints~(\ref{eq:predicate})--(\ref{eq:valency}), inference is reduced to computing the MSA on the contracted graph (solid arcs in the middle column).
    This may lead to solutions that do not satisfy constraints~(\ref{eq:predicate})--(\ref{eq:valency}) on the expanded graph (top example).
    However, the gradient of the smoothed constraint~(\ref{eq:valency}) will induce penalties (in red) to vertex and arc scores  that will encourage the \texttt{loc\_1} predicate to either be dropped from the solution or to have an outgoing arc to a \texttt{state\_all} argument.
    Computing the MSA on the contracted graph with penalties results in a solution that satisfies constraints~(\ref{eq:predicate})--(\ref{eq:valency}) (bottom example).
    }}
    \label{fig:pipeline}
\end{figure*}
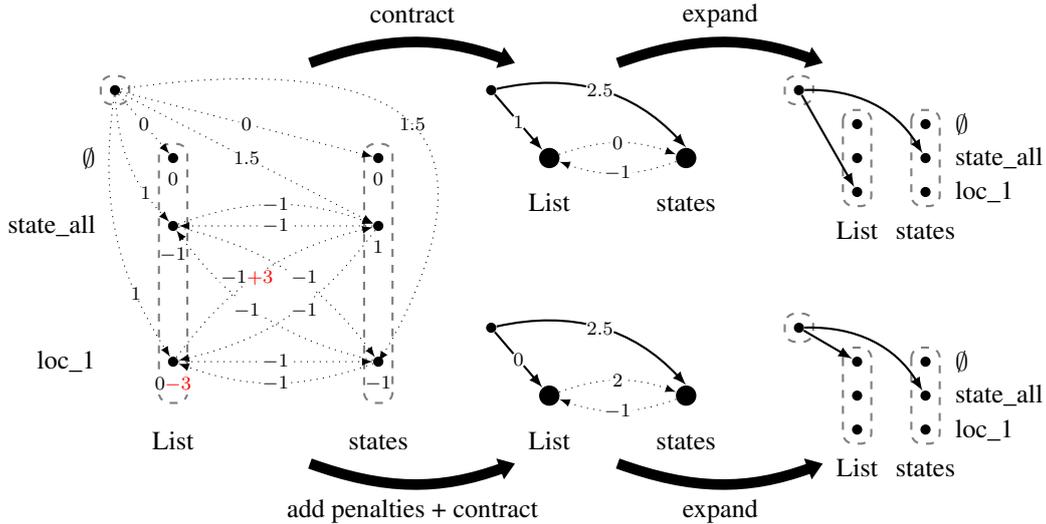

In this section, we propose an efficient way to solve the linear relaxations of MAP inference \textsc{(Ilp1)} and latent anchoring inference \textsc{(Ilp2)} via constraint smoothing and the conditional gradient method.
We focus on problems of the following form:
\begin{align*}
    \max_{\vz} \quad &f(\vz) \\
    \text{s.t.}\quad & \vz \in \conv(\mathcal{C}^{\text{(easy)}}) \\
    & \mA\vz = \vb \quad\text{or}\quad \mA\vz \leq \vb
\end{align*}
where
{\color{black}
the vector $\vz$ is the concatenation of the vectors $\vx$ and $\vy$ defined previously and
}
$\conv$ denotes the convex hull of a set. 
{\color{black}
We explained previously that if the set of constraints of form $\mA\vz = \vb$ for \textsc{(Ilp1)} or $\mA\vz \leq \vb$ for \textsc{(Ilp2)} was absent, the problem would be easy to solve under a linear objective function.
In fact, there exists an efficient linear maximization oracle (LMO), \emph{i.e.}\ a function that returns the optimal integral solution, for the set $\conv(\mathcal{C}^{\text{(easy)}})$.}
This setting covers both $\textsc{(Ilp1)}$ and $\textsc{(Ilp2)}$ where we have $\mathcal{C}^{\text{(easy)}} = \mathcal{C}^{\text{(sa)}}$ and $\mathcal{C}^{\text{(easy)}} = \mathcal{C}^{*\text{(relaxed)}}$, respectively.

An appealing approach in this setting is to introduce the problematic constraints as penalties in the objective:
\begin{align*}
    \max_{\vz} \quad& f(\vz) - \indicator_S(\mA\vz) \\
    \text{s.t.}\quad & \vz \in \conv(\mathcal{C}^{\text{(easy)}})
\end{align*}
where $\indicator_S$ is the indicator function of the set $S$:
\begin{align*}
    \indicator_S(\mA\vz) = \begin{cases}
        0 \quad&\text{if } \mA\vz \in S, \\
        +\infty \quad&\text{otherwise.}
    \end{cases}
\end{align*}
In the equality case, we use $S = \{ \vb \}$ and in the inequality case, we use $S = \{\vu | \vu \leq \vb \}$.

\subsection{Conditional gradient method}
Given a proper, smooth and differentiable function $g$ and a nonempty, bounded, closed and convex set $\conv(\mathcal{C}^{\text{(easy)}})$, the conditional gradient method \cite[a.k.a.\ Frank-Wolfe,][]{frank-wolfe1956,levitin1966constrained,lacoste2015fw} can be used to solve optimization problems of the following form:
\begin{align*}
\max_{\vz} \quad g(\vz) \quad\text{s.t.}\quad \vz \in \conv(\mathcal{C}^{\text{(easy)}})
\end{align*}
Contrary to the projected gradient method, this approach does not require to compute projections onto the feasible set $\conv(\mathcal{C}^{\text{(easy)}})$ which is, in most cases, computationally expensive.
Instead, the conditional gradient method only relies on a LMO:
$$
\lmo_{\mathcal{C}^{\text{(easy)}}}(\vpsi) \in \argmax_{\vz \in \conv(\mathcal{C}^{\text{(easy)}})} \vpsi^\top \vz.
$$
{\color{black}The algorithm constructs a solution to the original problem as a convex combination of elements returned by the LMO.}
The pseudo-code is given in Algorithm~\ref{alg:fw}.
An interesting property of this method is that its step size range is bounded.
This allows for simple linesearch techniques.

\subsection{Smoothing}

Unfortunately, the function $g(\vz) = f(\vz) - \indicator_S(\mA\vz)$ is non-smooth due to the indicator function term, preventing the use of the conditional gradient method.
We propose to rely on the framework proposed by \citet{pmlr-v80-yurtsever18a} where the indicator function is replaced by a smooth approximation.
The indicator function of the set $S$ can be rewritten as:
\begin{align*}
    \indicator_S(\mA\vz) = \indicator^{**}_S(\mA\vz) &= \sup_{\vu} \vu^\top \mA\vz - \support_S(\vu),
\end{align*}
where $\indicator^{**}_S$ denotes the Fenchel biconjugate of the indicator function and $\support_S(\vu) = \sup_{\vt \in S} \vu^\top \vt$ is the support function of $S$.
More details can be found in \citet[][Section 4.1 and 4.2]{beck2017first}.
In order to smooth the indicator function, we add a $\beta$-parameterized convex regularizer $-\frac{\beta}{2} \| \cdot \|_2^2$ to its Fenchel biconjugate:
$$
\indicator_{S, \beta}^{**}(\mA\vz) = \max_{\vu} \vu^\top \mA\vz - \support_S(\vu) - \frac{\beta}{2} \|\vu \|_2^2
$$
where $\beta > 0$ controls the quality and the smoothness of the approximation \cite{nesterov2005smooth}.

\textbf{Equalities.}
In the case where $S = \{ \vb \}$, 
with a few computations that are detailed by \citet{pmlr-v80-yurtsever18a}, we obtain:
$$
    \indicator_{\{ \vb \}, \beta}^{**}(\mA\vz) = \frac{1}{2 \beta}\|\mA \vz - \vb \|^2_2.
$$
That is, we have a quadratic penalty term in the objective.
Note that this term is similar to the term introduced in an augmented Lagrangian \cite[][Equation 17.36]{nocedal1999numerical},
and adds a penalty in the objective for vectors $\vz$ s.t.\ $\mA \vz \neq \vb$.

\textbf{Inequalities.}
In the case where $S = \{\vu | \vu \leq \vb \}$, 
similar computations lead to:
$$
    \indicator_{\leq \vb, \beta}^{**}(\mA\vz) = \frac{1}{2 \beta}\|[\mA \vz - \vb]_+ \|^2_2
$$
where $[\cdot]_+$ denotes the Euclidian projection into the non-negative orthant {\color{black}(\emph{i.e.}\ clipping negative values)}.
{\color{black}Similarly to the equality case, this term introduces a penalty in the objective for vectors $\vz$ s.t.\ $\mA \vz > \vb$.}
This penalty function is also called the Courant-Beltrami penalty function.

Figure~\ref{fig:pipeline} (bottom) illustrates how the gradient of the penalty term can ``force'' the LMO to return solutions that satisfy the smoothed constraints.

\subsection{Practical details}
\label{sec:practicalopt}
\textbf{Smoothness.}
In practice, we need to choose the smoothness parameter $\beta$.
We follow \citet{pmlr-v80-yurtsever18a} and use $\beta^{(k)} = \frac{\beta^{(0)}}{\sqrt{k + 1}}$ where $k$ is the iteration number and $\beta^{(0)} = 1$.

\textbf{Step size.}
Another important choice in the algorithm is the step size $\gamma$.
We show that when the smoothed constraints are equalities, computing the optimal step size has a simple closed form solution if the function $f$ is linear, which is the case for \textsc{(Ilp1)}, \emph{i.e.}\ MAP decoding.
The step size problem formulation at iteration $k$ is defined as:
\begin{align*}
\argmax_{\gamma \in [0, 1]} f(\vz^{(k)} + \gamma \vd) - \frac{\| \mA(\vz^{(k)} + \gamma \vd) - \vb \|^2}{2 \beta}
\end{align*}
By assumption, $f$ is linear and can be written as $f(\vz) = \vtheta^\top \vz$.
Ignoring the box constraints on $\gamma$, by first order optimality conditions, we have:
\begin{align*}
\gamma  &= \frac{-\beta\theta^\top \vd  + (\mA \vd)^\top\vb - (\mA \vd)^\top (\mA\vz^{(k)}) }{\|\mA \vd \|^2}
\end{align*}
We can then simply clip the result so that it satisfies the box constraints.
Unfortunately, in the inequalities case, there is no simple closed form solution.
We approximate the step size using 10 iterations of the bisection algorithm for root finding.

\textbf{Non-integral solutions.}
As we solve the linear relaxation of original ILPs, the optimal solutions may not be integral. 
Therefore, we use simple heuristics to construct a feasible solution to the original ILP in these cases.
For MAP inference, we simply solve the ILP\footnote{We use the multi-commodity flow formulation of \citet{martins-etal-2009-concise} instead of the cycle breaking constraints (\ref{eq:sa_connectivity}).} using CPLEX but introducing only variables that have a non-null value in the linear relaxation, leading to a very sparse problem which is fast to solve.
For latent anchoring, we simply use the Kuhn–Munkres algorithm using the non-integral solution as assignment costs.

\section{Experiments}
\begin{table*}
\small
\centering
\begin{tabular}{@{}l@{\hskip 0.25in}ccc@{\hskip 0.25in}ccc@{\hskip 0.25in}cc@{}}
\toprule
& \multicolumn{3}{c@{\hskip 0.25in}}{\scan} & \multicolumn{3}{c@{\hskip 0.25in}}{\geoquery} & \multicolumn{2}{c@{}}{\clevr}  \\
\cmidrule(r{0.25in}){2-4}\cmidrule(r{0.25in}){5-7}\cmidrule{8-9}
& \iid & \rright & \aright & \iid & \template & \length & \iid & \closure \\
\midrule
\multicolumn{9}{@{}l@{}}{\textbf{Baselines (denotation accuracy only)}} \\
\midrule
\textsc{Seq2Seq}
& 99.9 & 11.6 & 0
& 78.5 & 46.0 & 24.3
& \textbf{100} & 59.5
\\
\textsc{+ ELMo}
& \textbf{100} & 54.9 & 41.6
& 79.3 & 50.0 & 25.7
& \textbf{100} & 64.2
\\
\textsc{BERT2Seq}
& \textbf{100} & 77.7 & 95.3
& 81.1 & 49.6 & 26.1
& \textbf{100} & 56.4
\\
\textsc{GRAMMAR}
& \textbf{100} & 0.0 & 4.2
& 72.1 & 54.0 & 24.6
& \textbf{100} & 51.3
\\
\textsc{BART}
& \textbf{100} & 50.5 & \textbf{100}
& 87.1 & 67.0 & 19.3
& \textbf{100} & 51.5
\\
\textsc{SpanBasedSP}
& \textbf{100} & \textbf{100} & \textbf{100}
& 86.1 & 82.2 & 63.6
& 96.7 & 98.8
\\
\midrule
\multicolumn{9}{@{}l@{}}{\textbf{Our approach}} \\
\midrule
\color{black}Denotation accuracy
\color{black}& \color{black}\textbf{100} & \color{black}\textbf{100} & \color{black}\textbf{100}
\color{black}& \color{black}\textbf{92.9} & \color{black}\textbf{89.9} & \color{black}\textbf{74.9}
\color{black}& \color{black}\textbf{100} & \color{black}\textbf{99.6} \\
\color{black}\reflectbox{\rotatebox[origin=c]{180}{$\Rsh$}} Corrected executor
\color{black}& \color{black} & \color{black} & \color{black}
\color{black}& \color{black}91.8 & \color{black}88.7 & \color{black}74.5
\color{black}& \color{black} & \color{black}\\
\color{black}Exact match
& \textbf{100} & \textbf{100} & \textbf{100}
& 90.7 & 86.2 & 69.3
& \textbf{100} & \textbf{99.6} \\
\reflectbox{\rotatebox[origin=c]{180}{$\Rsh$}} w/o CPLEX heuristic
& \textbf{100} & \textbf{100} & \textbf{100}
& 90.0 & 83.0 & 67.5
& \textbf{100} & 98.0 \\
\bottomrule
\end{tabular}
\caption{Denotation and exact match accuracy on the test sets. All the baseline results were taken from \citet{herzig-berant-2021-span}.
{\color{black} For our approach, we also report exact match accuracy, \emph{i.e.} the percentage of sentences for which the prediction is identical to the gold program.
The last line reports the exact match accuracy without the use of CPLEX to round non integral solutions (Section~\ref{sec:practicalopt}).
}}
\label{table:results}
\end{table*}

We compare our method to baseline systems both on i.i.d.\ splits (\iid{}) and splits that test for compositional generalization for three datasets.
The neural network is described in Appendix~\ref{app:exp}.

\textbf{Datasets.}
\scan{} \cite{pmlr-v80-lake18a} contains natural language navigation commands. 
We use the variant of \citet{herzig-berant-2021-span} for semantic parsing.
The \iid{} split is the \emph{simple} split \cite{pmlr-v80-lake18a}. 
The compositional splits are \emph{primitive right} (\rright{}) and \emph{primitive around right} (\aright{}) \cite{loula-etal-2018-rearranging}.

\geoquery{} \cite{10.5555/1864519.1864543} uses the FunQL formalism \cite{10.5555/1619499.1619504} and  contains questions about the US geography.
The \iid{} split is the standard split and compositional generalization is evaluated on two splits: \length{} where the examples are split by program length and \template{} \cite{finegan-dollak-etal-2018-improving} where they are split such that all semantic programs having the same AST are in the same split.

\clevr{} \cite{Johnson2017CLEVRAD} contains synthetic questions over object relations in images. \closure{} \cite{DBLP:journals/corr/abs-1912-05783} introduces additional question templates that require compositional generalization.
We use the original split as our \iid{} split and the \closure{} split as a compositional split where the model is evaluated on \closure{}.

\textbf{Baselines.}
We compare our approach against the architecture proposed by \citet{herzig-berant-2021-span} (\textsc{SpanBasedSP}) as well as the seq2seq baselines they used.
In \textsc{Seq2Seq} \cite{jia-liang-2016-data}, the encoder is a bi-LSTM over pre-trained GloVe embeddings \cite{pennington-etal-2014-glove} or \textsc{ELMo} \cite{peters-etal-2018-deep} and the decoder is an attention-based LSTM \cite{DBLP:journals/corr/BahdanauCB14}.
\textsc{BERT2Seq} replaces the encoder with {\tt BERT-base}.
\textsc{GRAMMAR} is similar to \textsc{Seq2Seq} but the decoding is constrained by a grammar.
\textsc{BART} \cite{lewis-etal-2020-bart} is pre-trained as a denoising autoencoder.

\textbf{Results.}
We report the denotation accuracies in Table \ref{table:results}. 
Our approach outperforms all other methods.
In particular, the seq2seq baselines suffer from a significant drop in accuracy on splits that require compositional generalization.
While \textsc{SpanBasedSP} is able to generalize, our approach outperforms it.
Note that we observed that the \geoquery{} execution script used to compute denotation accuracy in previous work contains several bugs that overestimate the true accuracy.
Therefore, we also report denotation accuracy with a corrected executor\footnote{\url{ https://github.com/alban-petit/geoquery-funql-executor}} (see Appendix~\ref{sec:geo_exec}) for fair comparison with future work.

We also report exact match accuracy, with and without the heuristic to construct integral solutions from fractional ones.
The exact match accuracy is always lower or equal to the denotation accuracy.
This shows that our approach can sometimes provide the correct denotation even though the prediction is different from the gold semantic program.
Importantly, while our approach outperforms baselines, its accuracy is still significantly worse on the split that requires to generalize to longer programs.

\section{Related work}

\textbf{Graph-based methods.}
Graph-based methods have been popularized by syntactic dependency parsing \cite{mcdonald2005msa} where MAP inference is realized via the maximum spanning arborescence algorithm \cite{chu1965msa,edmonds1967msa}.
A benefit of this algorithm is that it has a $\mathcal O(n^2)$ time-complexity \cite{tarjan1977msa}, \emph{i.e.}\ it is more efficient than algorithms exploring more restricted search spaces \cite{eisner1997bilexical,gomez2011mildly,pitler2012gapminding,pitler2013endpoint}.

In the case of semantic structures, \citet{kuhlmann2015noncrossing} proposed a $\mathcal O(n^3)$ algorithm for the maximum non-necessarily-spanning acyclic graphs with a noncrossing arc constraint.
Without the noncrossing constraint, the problem is known to be NP-hard \cite{grotschel1985acyclic}.
To bypass this computational complexity, \citet{dozat2018semantic} proposed to handle each dependency as an independent binary classification problem, that is they do not enforce any constraint on the output structure.
Note that, contrary to our work, these approaches allow for reentrancy but do not enforce well-formedness of the output with respect to the semantic grammar.
\citet{lyu2018amr} use a similar approach for AMR parsing where tags are predicted first, followed by arc predictions and finally heuristics are used to ensure the output graph is valid.
On the contrary, we do not use a pipeline and we focus on joint decoding where validity of the output is directly encoded in the search space.

Previous work in the literature has also considered reduction to graph-based methods for other problems, \emph{e.g.}\ for discontinuous constituency parsing \cite{fernandez-gonzalez-martins-2015-parsing,corro-etal-2017-efficient}, lexical segmentation \cite{constant2015dependency} and machine translation \cite{zaslavskiy2009tsp}, \emph{inter alia}.

\textbf{Compositional generalization.}
Several authors observed that compositional generalization insufficiency is an important source of error for semantic parsers, especially ones based on seq2seq architectures \cite{pmlr-v80-lake18a,finegan2018text2sql,herzig2019detect,keysers2020measuring}.
\citet{NEURIPS2021_6f46dd17} proposed a latent re-ordering step to improve compositional generalization, whereas \citet{zheng-lapata-2021-compositional-generalization} relied on latent predicate tagging in the encoder.
There has also been an interest in using data augmentation methods to improve generalization \cite{jia-liang-2016-data, andreas-2020-good, akyurek2021learning, qiu-etal-2022-improving, yang-etal-2022-subs}.

Recently, \citet{herzig-berant-2021-span} showed that span-based parsers do not exhibit such problematic behavior.
Unfortunately, these parsers fail to cover the set of semantic structures observed in English treebanks, and we hypothesize that this would be even worse for free word order languages.
Our graph-based approach does not exhibit this downside.
Previous work by \citet{jambor-bahdanau-2022-lagr} also considered graph-based methods for compositional generalization, but their approach predicts each part independently without any well-formedness or acyclicity constraint.
\section{Conclusion}

In this work, we focused on graph-based semantic parsing for formalisms that do not allow reentrancy.
We conducted a complexity study of two inference problems that appear in this setting.
We proposed ILP formulations of these problems together with a solver for their linear relaxation based on the conditional gradient method.
Experimentally, our approach outperforms comparable baselines.

One downside of our semantic parser is speed (we parse approximately 5 sentences per second for \geoquery).
However, we hope this work will give a better understanding of the semantic parsing problem together with baseline for faster methods.

{\color{black}Future research will investigate extensions for (1) ASTs that contain reentrancies and (2) prediction algorithms for the case where a single word can be the anchor of more than one predicate or entity.
These two properties are crucial for semantic representations like Abstract Meaning Representation \cite{banarescu2013amr}.
Moreover, even if our graph-based semantic parser provides better results than previous work on length generalization, this setting is still difficult.
A more general research direction on neural architectures that generalize better to longer sentences is important.
}

\section*{Acknowledgments}

We thank François Yvon and the anonymous reviewers for their comments and suggestions.
We thank Jonathan Herzig and Jonathan Berant for fruitful discussions.
This work benefited from computations done on the Saclay-IA platform and on the HPC resources of IDRIS under the allocation 2022-AD011013727 made by GENCI.

\bibliography{tacl2021}
\bibliographystyle{acl_natbib}

\appendix
\clearpage
\section{Proofs}
\label{proofs}


\begin{proof}[Proof: Theorem~\ref{th:parsing}]
We prove Theorem~\ref{th:parsing} by reducing the maximum not-necessarily-spanning arborescence (MNNSA) problem, which is known to be NP-hard \cite{rao2002nnsa,duhamel-etal-2008-models}, to the MGVCNNSA.

Let $G = \langle V, A, \vpsi \rangle$ be a weighted graph where $V = \{0, ..., n \}$ and $\vpsi \in \R^{|A|}$ are arc weights.
The MNNSA problem aims to compute the subset of arcs $B \subseteq A$ such that $\langle V[B], B \rangle$ is an arborescence of maximum weight, where its weight is defined as $\sum_{a \in B} \evpsi_a$.

Let $\lang = \langle \tags, \types, \ftype, \fargs \rangle$ be a grammar such that $\tags = \{0, ..., n-1 \}$, $\types = \{ t \}$ and $\forall e \in E: \ftype(e) = t \wedge \fargs(e, t) = e$.
Intuitively, a tag $e \in E$ will be associated to vertices that require exactly $e$ outgoing arcs.

We construct a clustered labeled weighted graph $G' = \langle V', A', \pi, \flabeling, \vpsi' \rangle$ as follows. 
$\pi = \{ V'_0, ..., V'_n \}$ is a partition of $V'$ such that each cluster $V'_i$ contains $n-1$ vertices and represents the vertex $i \in V$.
The labeling function $\flabeling$ assigns a different tag to each vertex in a cluster, \emph{i.e.} $\forall V'_i \in \pi, \forall u', v' \in V'_i: u' \neq v' \Rightarrow l(u') \neq l(v')$.
The set of arcs is defined as $A' = \{ u' \to v' | \exists i \to j \in A \text{ s.t.\ } u' \in V'_i \land v' \in V'_j \}$.
The weight vector $\vpsi' \in \R^{|A'|}$ is such that $\forall u' \to v' \in A' : u' \in V'_u \wedge v' \in V'_v \Rightarrow \evpsi'_{u' \to v'} = \evpsi_{u \to v}$.

As such, there is a one-to-one correspondence between solutions of the MNNSA on graph $G$ and solutions of the MGVCNNSA on graph $G'$.
\end{proof}

Note that our proof considers that arcs leaving from the root cluster satisfy constraints defined by the grammar whereas we previously only required the root vertex to have a single outgoing arc.
The latter constraint can be added directly in a grammar, but we omit presentation for brevity.
{\color{black}
The constrained arity case presented by \citet{mcdonald2007complexity} focuses on spanning arborescences with an arity constraint by reducing the Hamiltonian path problem to their problem.
While the arity constraint is similar in their problem and ours, our proof considers the not-necessarily-spanning case instead of the spanning one.
Although the two problems seem related, they need to be studied separately, \emph{e.g.} computing the maximum spanning arborescence is a polynomial time problem whereas computing the MNNSA is a NP-hard problem.}

\begin{proof}[Proof: Theorem~\ref{th:assignment}]
We prove Theorem~\ref{th:assignment} by reducing the maximum directed Hamiltonian path problem, which is known to be NP-hard \cite[][Appendix A1.3]{garey1979computers}, to the latent anchoring.

Let $G = \langle V, A, \vpsi \rangle$ be a weighted graph where $V = \{1, ..., n\}$ and $\vpsi \in \R^{|A|}$ are arc weights.
The maximum Hamiltonian path problem aims to compute the subset of arcs $B \subseteq A$ such that $V[B] = V$ and $\langle V[B], B \rangle$ is a path of maximum weight, where its weight is defined as $\sum_{a \in B} \evpsi_a$.

Let $\lang = \langle \tags, \types, \ftype, \fargs \rangle$ be a grammar such that $\tags = \{0, 1\}$, $\types = \{ t \}$ and $\forall e \in E: \ftype(e) = t \land \fargs(e, t) = e$.

We construct a clustered labeled weighted graph $G' = \langle V', A', \pi, \flabeling, \vpsi' \rangle$ as follows. 
$\pi = \{ V'_0, ..., V'_n \}$ is a partition of $V'$ such that $V'_0 = \{ 0 \}$, each cluster $V'_i \neq V'_0$ contains 2 vertices and represents the vertex $i \in V$.
The labeling function $\flabeling$ assigns a different tag to each vertex in a cluster except the root, \emph{i.e.} $\forall V'_i \in \pi, i > 0, \forall u', v' \in V'_i: u' \neq v' \Rightarrow l(u') \neq l(v')$.
The set of arcs is defined as $A' = \{0 \to u' | u' \in V' \setminus \{0\}\} \cup \{ u' \to v' | i \to j \in A \land u' \in V'_i \land v' \in V'_j \}$.
The weight vector $\vpsi' \in \R^{|A'|}$ is such that $\forall u' \to v' \in A' : u' \in V'_u \wedge v' \in V'_v \Rightarrow \evpsi'_{u' \to v'} = \evpsi_{u \to v}$ and arcs leaving $0$ have null weights.

We construct an AST $G'' = \langle V'', A'', \flabeling' \rangle$ such that $V'' = \{1, ..., n\}$, $A'' = \{ i \rightarrow i+1 ~ | ~ 1 \leq i < n \}$ and the labeling function $\flabeling'$ assigns the tag $0$ to $n$ and the tag $1$ to every other vertex.

As such, there is a one-to-one correspondence between solutions of the maximum Hamiltonian path problem on graph $G$ and solutions of the mapping of maximum weight of $G''$ with $G'$.
\end{proof}
\section{Experimental setup}
\label{app:exp}

The neural architecture used in our experiments to produce the weights $\vmu$ and $\vphi$ is composed of:
(1) an embedding layer of dimension 100 for \scan{} or {\tt BERT-base} \cite{devlin-etal-2019-bert} for the other datasets, followed by a bi-LSTM \cite{hochreiter1997lstm} with a hidden size of 400;
(2) a linear projection of dimension 500 over the output of the bi-LSTM followed by a \textsc{Tanh} activation and another linear projection of dimension $|E|$ 
to obtain $\vmu$; 
(3) a linear projection of dimension 500 followed by a \textsc{Tanh} activation and a bi-affine layer \cite{dozat2017deep} to obtain $\vphi$.

We apply dropout with a probability of 0.3 over the outputs of {\tt BERT-base} and the bi-LSTM and after both \textsc{Tanh} activations.
The learning rate is $5\times10^{-4}$ and each batch is composed of 30 examples.
We keep the parameters that obtain the best accuracy on the development set after 25 epochs.
Training the model takes between 40 minutes for \geoquery{} and 8 hours for \clevr{}. However, note that the bottleneck is the conditional gradient method which is computed on the CPU.
\section{\geoquery{} denotation accuracy issue}
\label{sec:geo_exec}

{\color{black}
The denotation accuracy is evaluated by checking whether the denotation returned by an executor is the same when given the gold semantic program and the prediction of the model.
It can be higher than the exact match accuracy when different semantic programs yield the same denotation.

When we evaluated our approach using the same executor as the baselines of \citet{herzig-berant-2021-span}, we observed two main issues regarding the behaviour of several predicates:
(1) Several predicates have undefined behaviours (\emph{e.g.}\ \texttt{population\_1} and \texttt{traverse\_2} in the case of an argument of type \texttt{country}), in the sense that they are not implemented;
(2) The behaviour of some predicates are incorrect with respect to their expected semantic (\emph{e.g.}\ \texttt{traverse\_1} and \texttt{traverse\_2}).
These two sources of errors result in incorrect denotation for several semantic programs, leading to an overestimation of the denotation accuracy when both the gold and predicted programs return by accident an empty denotation (potentially for different reasons, due to aforementioned implementation issues).

We implemented a corrected executor addressing the issues that we found.

}

\end{document}